\documentclass[manuscript, screen]{acmart}

\usepackage{multirow}

\AtBeginDocument{%
  }

\begin{document}

\title[AnnualBERT]{Towards understanding evolution of science through language model series}

\author{Junjie Dong}
\email{jd445@qq.com}
\orcid{0000-0001-8267-9181}
\affiliation{%
  \institution{City University of Hong Kong}
  \city{Hong Kong}
  \country{China}
}

\author{Zhuoqi Lyu}
\email{zqlyu2-c@my.cityu.edu.hk}
\orcid{0000-0002-5787-0862}
\affiliation{%
  \institution{City University of Hong Kong}
  \city{Hong Kong}
  \country{China}
}

\author{Qing Ke}
\email{q.ke@cityu.edu.hk}
\orcid{0000-0002-2945-5274}
\affiliation{%
  \institution{City University of Hong Kong}
  \city{Hong Kong}
  \country{China}
}

\begin{abstract}
We introduce AnnualBERT, a series of language models designed specifically to capture the temporal evolution of scientific text. Deviating from the prevailing paradigms of subword tokenizations and ``one model to rule them all'', AnnualBERT adopts whole words as tokens and is composed of a base RoBERTa model pretrained from scratch on the full-text of 1.7 million arXiv papers published until 2008 and a collection of progressively trained models on arXiv papers at an annual basis. We demonstrate the effectiveness of AnnualBERT models by showing that they not only have comparable performances in standard tasks but also achieve state-of-the-art performances on domain-specific NLP tasks as well as link prediction tasks in the arXiv citation network. Our approach enables the pretrained models to not only improve performances on scientific text processing tasks but also to provide insights into the development of scientific discourse over time. The series of the models is available at \url{https://huggingface.co/jd445/AnnualBERTs}.
\end{abstract}



\maketitle

\section{Introduction}\label{sec:introduction}

Natural language processing (NLP) has been experiencing a new paradigm where Transformers-based language models pretrained on a massive text corpus have significantly improved performance in downstream NLP tasks~\cite{vaswani2017attention}. In our focused scientific domains, both the encoder and decoder components of Transformers have been extensively employed to develop scientific language models pretrained on the full-text of research papers~\cite{ho2024survey}. 

When developing language models for a target domain, previous studies have suggested a number of pretraining methods. The first is continual training, which uses a model pretrained already on a general domain corpus and continues the pretraining process on the target domain corpus (\emph{i.e.}, domain adaption, mixed-domain pretraining)~\cite{gururangan2020dont}. BioBERT is one example adapted from the original BERT model to the biomedical domain using the PubMed corpus~\cite{lee2020biobert}. The second approach is training from scratch directly on the target domain corpus. Recent development of BERT-like models, including both multi-domain~\cite{liu2022oag, beltagy2019scibert} and domain-specific ones~\cite{gu2021domain, naseem2021bioalbert, trewartha2022quantifying}, adopts this strategy to better capture the linguistic and semantic features of their respective fields. Training from scratch has been shown to outperform continual training, because it constructs a vocabulary that encompasses more in-domain tokens, leading to more effective learning from the domain corpus. By contrast, continual training retains the original model's vocabulary and consequently fragments complex, domain-relevant words into less meaningful tokens. For example, the word ``chloramphenicol'' is treated as one token by BiomedBERT (PubmedBERT) but six tokens by BioBERT (``ch-lor-amp-hen-ico-l''). 

\begin{figure*}[t!]
\centering
\includegraphics[trim=0 0 0 0, width=1\linewidth]{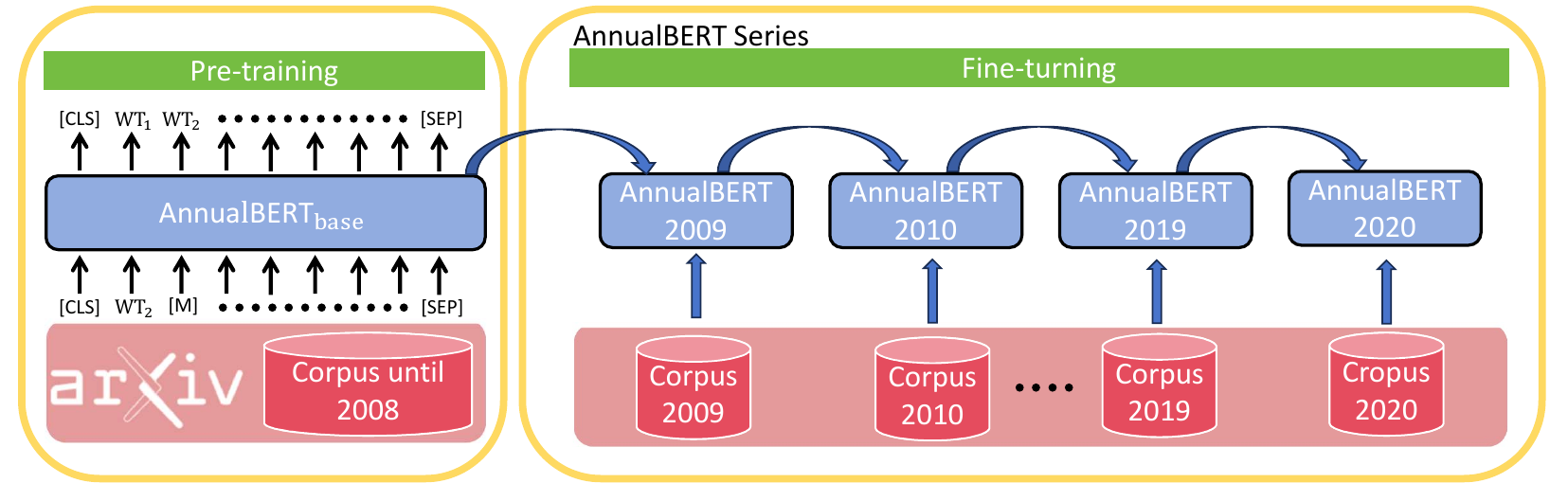}
\caption{Workflow for developing our AnnualBERT models pretrained on arXiv papers. We first train from scratch a RoBERTa model on papers published until 2008, denoted as $\mathcal{M}_{\text{base}}$, and then for each subsequent year $t$ from 2009, we use papers published in that year for continual training of $\mathcal{M}_{t-1}$ to obtain $\mathcal{M}_{t}$. During the training process, we use whole word token, represented as $\text{WT}_1$, $\text{WT}_2$, \dots, $\text{WT}_n$.}
\label{fig:AnnualBERT}
\end{figure*}

Here we propose a novel pretraining strategy that consists of two stages: domain specific pretraining from scratch followed by progressively continual training on corpora within the same domain organized by year (Fig.~\ref{fig:AnnualBERT}). Our method is designed to, in addition to recognizing the importance of domain specific pretraining, capture the temporal evolution of language use. Temporal change is one of the most salient features in real-world language data: Scientific papers are published in a chronological order, collectively representing the evolving landscape of scientific knowledge during its development; similarly, news articles and social media posts are produced over time, the collection of which shapes the global news agenda. However, existing efforts, regardless of the training methods, largely ignore the temporal aspect of language data and instead are directed toward developing ``one model to rule them all''---one model that can excel in various downstream scientific NLP tasks like named entity recognition. 

By comparison, our proposal enables us to develop a series of language models on scientific documents, each of which represents one historical period and the collection of which allows us to understand the evolution of science. At a broader level, our vision is that a language model itself can be regarded as a highly knowledgeable researcher with extensive reading capabilities across all the scientific fields, who updates its inner workings (\emph{i.e.}, model parameters) upon reading all the papers published in a year. By probing how different versions of the model perform on the same task and by mining how model parameters change, we may be able to uncover how science evolves temporally. From another perspective, language models serve to compress documents over a certain period, thereby reflecting various characteristics of language use during that time, such as vocabulary, semantics, and topics. In this regard, our work is related to some recent studies that identify the temporal misalignment phenomenon~\cite{rijhwani2020temporally, luu2022time} and the remedies to derive customized models to different periods~\cite{luu2022time, wang2023comprehensive, nylund2023time}. However, the language models used in the first place was pretrained on a time discorded corpus, which confounds the derivation. 

To demonstrate the utility of our method, we present AnnualBERT as a running example. It is a series of language models developed by first pretraining RoBERTa on a large corpus of full-text of arXiv papers published until 2008 and then continually training on the corpora formed by arXiv papers in individual later years. We conduct extensive experimentation to demonstrate the validity and utility of the AnnualBERT models. Specifically, the main contributions of our study are summarized as follows. 
\begin{itemize}
\item We propose a novel training method that involves organizing scientific text by year and continually training our language model accordingly. We release the pretrained AnnualBERT, a series of language models specifically tailored for scientific text. 

\item We evaluate AnnualBERT models on a suit of standard NLP tasks in the scientific domain, showing comparable performances to similar models. Moreover, we formulate several domain-specific tasks and demonstrate that AnnualBERT has better performances for these in-domain tasks. 

\item We showcase the utility of AnnualBERT models by performing systematic evaluations on link prediction tasks in the arXiv citation network. Under both static and temporal settings, AnnualBERT achieves uniquely better performances, indicating its effectiveness in generating more informative embeddings for the scientific domain. 

\item Through model weight visualizations and a series of probe tasks, we quantify the extent of representation learning and forgetting of AnnualBERT models and find that the learning and forgetting processes are highly task-specific. 
\end{itemize}

\section{Related work} \label{sec:related}

\subsection{Scientific language models} \label{subsec:scilm}

The Transformers architecture, along with the growing accessibility of full-text corpora of scientific papers, has heightened the significance of employing NLP techniques for document understanding and knowledge extraction~\cite{ho2024survey}. Similar to general domains, transfer learning through pretraining models can leverage information learned in a self-supervised fashion to aid in downstream scientific NLP tasks where datasets are scarce, leading to the introduction of numerous scientific language models (SciLMs) for a number of fields, including biomedicine \cite{beltagy2019scibert, lee2020biobert, naseem2021bioalbert, kanakarajan2021BioELECTRA, luo2022BioGPT, bolton2024biomedlm}, chemistry~\cite{guo2021automated}, material science~\cite{trewartha2022quantifying}, and social science~\cite{shen2023sscibert}. 

\begin{table*}[!t]
\caption{Different domain-specific language models for scientific text.}
\centering
\resizebox{\textwidth}{!}{
\begin{tabular}{l c c c c c c c c}
\toprule
{Model} & {Domain} & {Corpus} & {Tokenization} &{Vocab.}  & {Pretraining} & {Corpus Size}   \\
\midrule
BERT                & General & Wiki + Books    &  WordPiece & 28996 & Scratch & 3.3B words / 16GB    \\ 
BioBERT             & Bio   & PubMed abstract &  WordPiece & 28996 & Continual & 4.5B words \\ 
SciBERT             & Multi & Semantic Scholar full text &  WordPiece & 31090 & Scratch & 3.2B words \\ 
BiomedBERT          & Bio   & PubMed abstract &  WordPiece & 28895 & scratch & 3.2B words / 21GB  \\
ScholarBERT         & Multi & Public.Resource.org full text &  WordPiece & 50000 & Scratch & 221B words   \\ 
AnnualBERT$_{2008}$ & Multi & arXiv full text &  Whole word & 53072 & Scratch & 1.2B words / 12GB    \\ 
AnnualBERT$_{2020}$ & Multi & arXiv full text &  Whole word & 53072 & Scratch & 2.7B words / 41GB    \\ 
\bottomrule
\end{tabular}
}
\label{tab:bert-zoo}
\end{table*}

Table~\ref{tab:bert-zoo} summarizes some notable SciLMs that employ the BERT architecture. BioBERT~\cite{lee2020biobert} was resulted from continual pretraining of the original BERT model on PubMed abstracts and showed better performances on biomedical NLP tasks. 
SciBERT~\cite{beltagy2019scibert} was pretrained from scratch on the full-text of 1.1M biomedical and computer science papers in the Semantic Scholar corpus~\cite{s2orc}. Evaluations on multiple tasks showed over-performances than the original BERT and comparable performances as BioBERT on biomedical tasks. 
BiomedBERT~\cite{gu2021domain} was also a pretrained model from scratch on 3.2B tokens in 14M PubMed abstracts. The work highlights the necessity of pretraining from scratch, as it yields a domain specific vocabulary that is useful for downstream tasks. In their case, the vocabulary contains more biomedical terms like ``acetyltransferase'', in contrast to the vocabulary in BioBERT that consists of terms common to the general domain. 
OAG-BERT~\cite{liu2022oag} was trained on the AMiner and PubMed corpora. It is based on the BERT architecture but fine-tuned for academic graph related tasks such as author name disambiguation, paper recommendation, and citation prediction. 
ScholarBERT \cite{hong2023diminishing} was pretrained from scratch on perhaps the largest corpus (221B tokens in 75M journal articles provided by public.resource.org). One observation from this work is that performances in downstream scientific tasks saturate when increasing training data, model size, or training time. 

In addition to scientific text, other modalities of scientific data have been used for pretraining language models, including math equations~\cite{peng2021mathbert}, chemicals, proteins, etc. Notably, SMILES strings of chemicals have been utilized to pretrain BERT-like models, yielding several chemical language models~\cite{ross2022large, ahmad2022ChemBERTa2}. Similarly, protein language models are trained using protein sequences~\cite{elnaggar2021prottrans}. 

\subsection{Temporality of language models}

As we shall pretrain SciLMs on temporally organized corpora, our work is broadly related to training models adaptive to diverse domains (\emph{e.g.}, \cite{huang2024self}, \cite{fan2024generalizing}) and specifically related to the temporal aspect of language models. Some studies found that language models implicitly encode various external knowledge including space and time~\cite{gurnee2023language}. However, an increasing number of studies reported the temporal misalignment phenomenon, which means that models trained on text from one period can have performance degradation when tested on text from another. For example, \cite{rijhwani2020temporally} showed the effect of temporal drift on NER tasks in temporally-diverse tweets. Likewise, \cite{luu2022time} quantified performance degradation due to temporal misalignment across domains and tasks. These studies corroborate the importance of customizing models to different time periods, which is the practice we adopt in this work. To mitigate the effect of temporal misalignment, existing studies suggested a few methods, including continuous pretraining~\cite{luu2022time, wang2023comprehensive}, adding year flags to training instances~\cite{dhingra2022time}, and discarding outdated facts~\cite{zhang2023mitigating}. However, these methods are proposed to improve performances in downstream tasks, but it turned out that the improvement is limited~\cite{luu2022time}. 

Another strategy aims at achieving a desired outcome by directly editing a model itself, which involves altering a pretrained model in the weight space to enhance model performance in specific tasks~\cite{ilharco2022editing, nylund2023time}. Current approaches to model editing focused on task arithmetic~\cite{ilharco2022editing} and weight interpolation~\cite{matena2022merging, wortsman2022model}. In the context of addressing temporal misalignment, model editing through time vector as a special case of task vector was used to obtain an updated pretrained model for a specific period~\cite{nylund2023time}. However, a critical limitation of time vector based model editing is that the language model used in the first place was pretrained on a time discorded corpus, which confounds the derivation of time vectors. Moreover, using such language model in the scientific domain may be especially problematic, as the corpus size grows drastically due to the exponential growth of science publishing, which means that later corpora would over-shadow earlier ones in determining model weights. By contrast, instead of using model editing, we obtain an updated model through continual training on a series of corpora organized temporally. 

In this regard, our pretraining strategy can be considered as a special case of domain adaption~\cite{gururangan2020dont}, where each corpus in a separate period is treated as a ``domain''. A natural consequence of domain adaption that was studied is representation forgetting: As a language model adapts to new domains, it forgets some knowledge learned from previous ones, leading to performance degradation~\cite{lazaridou2021mind}. However, in our work, we find that representation forgetting is highly dependent on tasks: We identify two rather similar tasks where for one of them, representation forgetting does manifest, but for the other the opposite of forgetting is observed. 

\subsection{Computational analysis of evolution of science}

Distinguishing from the majority of existing SciLMs that concern about downstream NLP tasks, our motivation of pretraining a series of SciLMs is to utilize them to understand quantitatively how science evolves over time through the lens of scientific text. Thus, our work is related to inquiry into the evolution of science, using diverse types of data including full-text and metadata like authorship and citations. \cite{blei2006dynamic} proposed dynamic topic models and applied them to analyze the temporal evolution of topics in papers published in the journal \emph{Science}. Analyzing authors' transitions between different topics over time, \cite{anderson2012towards} identified distinctive stages in the history of computational linguistics. \cite{prabhakaran2016predicting} developed a classifier to assign rhetorical functions (\emph{e.g.}, background, method, conclusion, etc.) to sentences in abstracts and demonstrated the ability to predict the rise and fall of scientific topics by tracking the trajectories of rhetorical functions appeared in different topics over time. \cite{jurgens2018measuring} developed a classifier to label citation functions (\emph{e.g.}, background, motivation, etc.) and applied it to NLP papers to reveal the evolution of this field.

\section{Developing AnnualBERT}

\subsection{Preparing corpus}

As the number of scientific papers has surged, researchers have increasingly opted to post their manuscripts on preprint platforms to share with the scientific community. arXiv is perhaps the first such platform, which was initially for physics and over time expanded to many other fields. It has gained significant popularity recently and archived nearly 2.4 million papers. 

\begin{figure*}[t!]
\centering
\includegraphics[trim=0 8mm 0 0, width=\linewidth]{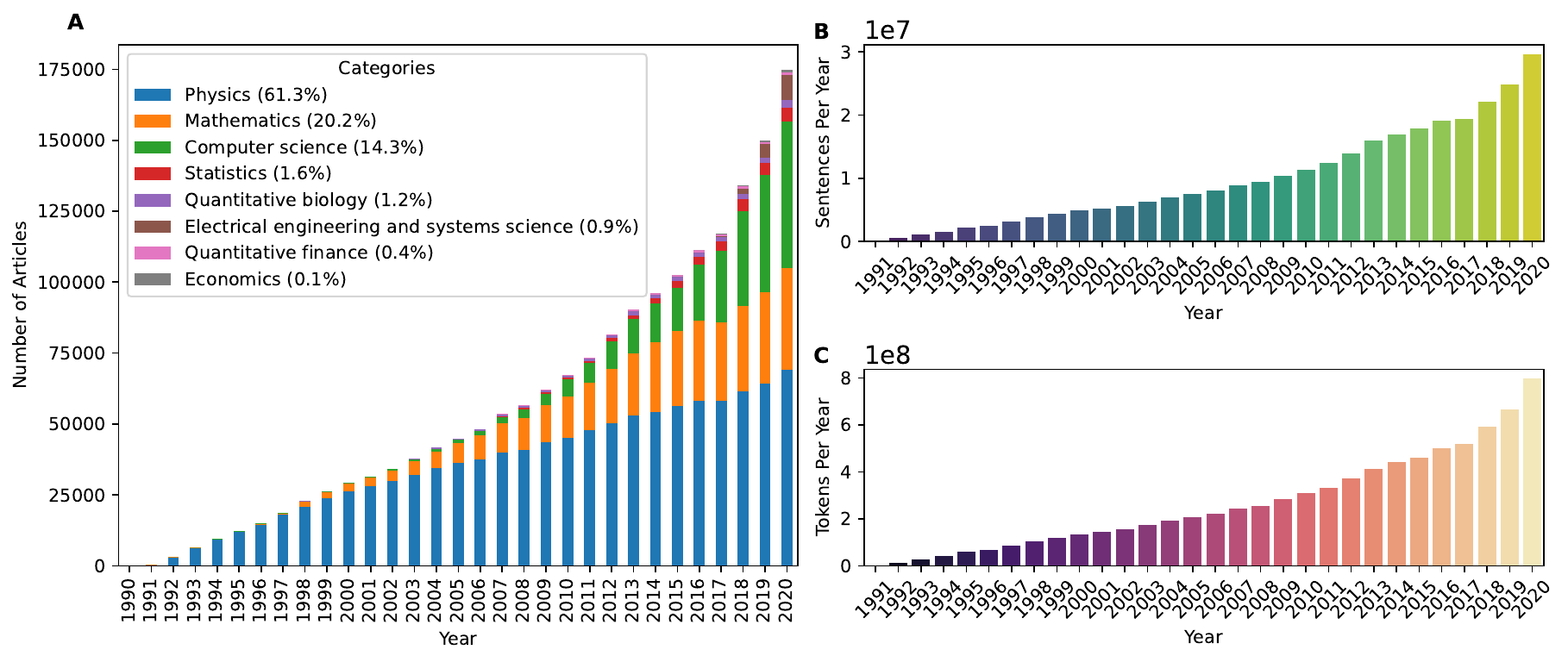}
\caption{Summary statistics of our arXiv corpus. (a) Yearly number of papers by category. (b) Yearly number of sentences and tokens after cleaning.}
\label{fig:stats}
\end{figure*}

We download from the arXiv website the source files of all the $1,752,637$ papers posted between 1990 and 2020. Fig.~\ref{fig:stats}a shows the field decomposition of papers by year, indicating that physics constantly dominates the corpus. In total, it accounts for 61.3\% of the articles, followed by mathematics (20.2\%). Computer science publications, although represent only 14.3\%, have increased exponentially in recent years. Papers from the other fields---Quantitative Biology, Quantitative Finance, Statistics, Electrical Engineering and Systems Science, and Economics---all together constitute less than 5\% of the dataset. 

To pretrain our models, we first need to prepare a full-text corpus extracted from the downloaded files, which are overwhelmingly \LaTeX{} and PDF files. For \LaTeX{} files, we use the \texttt{LaTeXML} tool~\cite{LaTeXML} to convert them into HTML files, which facilitates our followup processing, as there are myriad author-defined commands and symbols in the \LaTeX{} files. We then remove contents like figures, tables, and equations as well as metadata like author information, acknowledgment, and pagination, by striping away HTML elements corresponding to these contents. Next we replace certain markups with particular tokens, \emph{e.g.}, \texttt{[EQU]} for inline equations, \texttt{[CITE]} for citations, etc., which improves the clarity and utility of the text for training. As for PDF documents, we first use \texttt{PyMuPDF} to extract plain text and employ \texttt{neatext} to further clean elements like tables, emojis, and URLs, considering that PDF files often lack a clear document structure that is conducive to cleaning. We then filter out sentences with less than 20 characters and sentences that contain more than 40\% of non-essential elements like punctuation, special characters, stopwords, emojis, and URLs. The code for data processing and training can be found at \url{https://github.com/jd445/TrainAnuualBERT}. 

After extracting text from \LaTeX{} and PDF files, we use \texttt{NLTK}~\cite{loper2002nltk} for sentence segmentation and filter out sentences with fewer than three words. This completes our corpus preparation, and Fig.~\ref{fig:stats}b presents the total number of sentences over time.

\begin{figure}[t!]
\centering
\includegraphics[trim=0 5mm 0 0, width=0.5\linewidth]{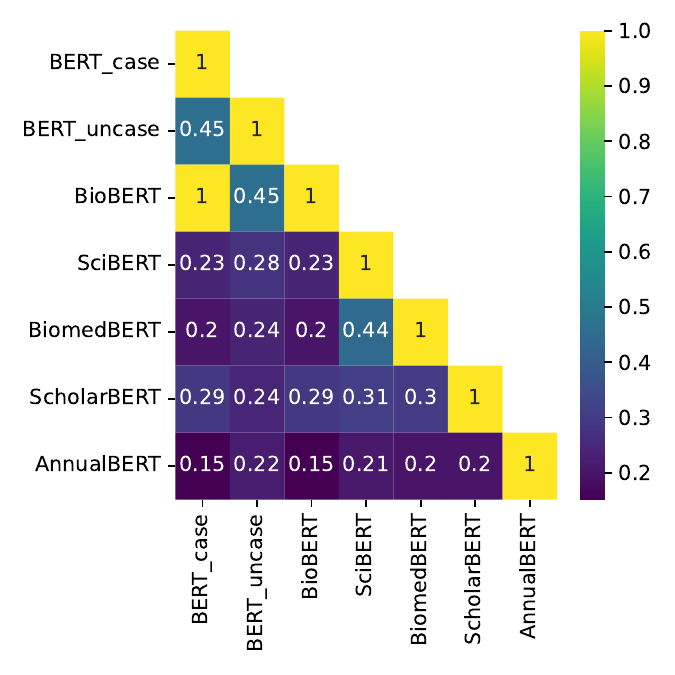}
\caption{Jaccard similarity matrix between vocabularies. }
\label{fig:vocab-sim}
\end{figure}

Next, we build our customized vocabulary. Existing language models use different tokenization methods to build vocabularies. The original BERT, for example, uses WordPiece, whereas RoBERTa uses BPE. Both tokenization methods consider subword as token, which has the advantage in representing unseen words by combining a list of existing tokens in the vocabulary. Subword tokenization, however, is not suitable for our purpose of examining temporal evolution of scientific discourse at the word level. Specifically, in \S~\ref{subsec:temp-adapt}, we investigate the development of scientific topics by tracking the changing probabilities of a word assigned by different versions of AnnualBERT, a task that is otherwise challenging to accomplish if that word is fragmented into several tokens. We thus utilize whole words as tokens, an approach scatteredly taken in previous studies~\cite{feng2024pretraining}. Particularly, we employ \texttt{spaCy} to tokenize all the abstracts from 1990 to 2020 into words and keep those with frequency $\geq 50$, resulting in a vocabulary with about $53,000$ words. We then add some special tokens including \texttt{[CITE]}, \texttt{[EQU]}, \texttt{[FIG]}, \texttt{[REF]}, and \texttt{[SEC]} to the vocabulary for ease of \LaTeX{} markup representation. This tailored approach ensures that our vocabulary is appropriate for interpretations of terminology pertaining to distinct fields of study. Note that words that appear in later corpora bear no influence on earlier models, as the corresponding entries in the embedding layers are kept unchanged as the initial values during model training. 

Fig.~\ref{fig:stats}b presents the total number of tokens over time. Fig.~\ref{fig:vocab-sim} presents the Jaccard similarity matrix between the vocabularies of different BERT models (see \S~\ref{subsec:scilm} for their details). We note that the similarities between BERT models and the other models (except BioBERT) are generally around 20\%. BioBERT and BERT\_{case} have similarity of one, since BioBERT adopts the vocabulary of BERT\_{case}. The similarities among SciBERT, BiomedBERT, and ScholarBERT are approximately 30\%, which is higher than that of BERT. This reflects differing vocabulary distributions between general and scientific domains. Our AnnualBERT exhibits lower similarities between the other models, which is attributed to the unique tokenization strategy. 

\subsection{Pretraining AnnualBERT}

The original BERT model is based on a bidirectional Transformer architecture and is pretrained through two objectives: Masked Language Model (MLM) and Next Sentence Prediction (NSP). In MLM, $15\%$ of the input tokens are masked, among which $80\%$ are replaced with the \texttt{[MASK]} token, $10\%$ are replaced with randomly selected tokens from the vocabulary, and the remaining $10\%$ remain unchanged. The objective is to predict the masked tokens using cross-entropy loss. The NSP task predicts whether two masked sentences follow one another in the original text. Building on BERT, the RoBERTa model~\cite{liu2019roberta} improves the pretraining by eliminating the NSP objective, which has been found less useful, and by implementing dynamic masking, where masked tokens change with each epoch, in contrast to the fixed masking used in BERT where masked tokens are fixed across epochs. 

Here we adopt the RoBERTa implementation to pretrain from scratch our base AnnualBERT model, $\mathcal{M}_{2008}$, on a corpus formed by all papers published up to 2008, which has about 1.2 billion words. The architecture of our AnnualBERT matches with the BERT$_\text{base}$ model, namely 12 hidden layers, 12 attention heads, and 768 hidden dimensions, totalling 110M parameters. For pretraining, we use the standard BERT tokenization approach while incorporating our own customized arXiv vocabulary. We set the vocabulary size to $53,100$, the \texttt{max\_len} parameter to $512$, and lowercase the corpus and the vocabulary. The pretraining process took around 108 hours to finish two epochs on a computing platform with 8 Nvidia A40 GPUs. 

With $\mathcal{M}_{2008}$, we perform, for each year $t > 2008$, continual training of $\mathcal{M}_{t-1}$ on the corpus formed by papers published in $t$ for one epoch, resulting in $\mathcal{M}_{t}$. For instance, a continual training of $\mathcal{M}_{2008}$ on the corpus based on papers published in 2009 yields $\mathcal{M}_{2009}$, which is then used to develop $\mathcal{M}_{2010}$ using papers in 2010, in general following: 
\begin{equation}
\mathcal{M}_{t} = \text{Continual-train} \left( \mathcal{M}_{t-1},\ \text{corpus}_{t} \right) \, .
\end{equation}
Through this approach, we allow each version of our AnnualBERT to capture evolving language and terminology specific to that year's publications, so that we can represent the knowledge of a specific year in a more sophisticated way. By the time we extend the model to AnnualBERT$_{2020}$, the corpora used for pretraining our AnnualBERT have grown to 2.7 billion words. 

Finally, to further understand the effect of our pretraining method, we separately pretrain another RoBERTa model, denoted as $\overline{\mathcal{M}}$, on the entire, globally shuffled corpus (all papers published until 2020) for one epoch. This pretraining is completed in about 7 days on a server with 4 Nvidia L40 GPUs.

\section{Performance on downstream NLP tasks}

\subsection{Task and dataset descriptions}

To validate our AnnualBERT models, we experiment on 3 downstream NLP tasks: named entity recognition (NER), text classification (CLS), and relation classification (REL). In designing those experiments, we consider tasks using benchmark datasets and tasks designed specifically for the arXiv domain. For benchmark datasets, many existing ones for scientific text are from biomedical domains. However, texts from those fields are largely absent from our training corpus. Therefore, in our experiment we do not include biomedical related benchmark datasets. Instead, we collect and categorize datasets from computer science and multi-subject domains, given that the arXiv corpus used in our work tends to come from science and engineering disciplines. Particularly, for NER tasks, we use the following 3 datasets. 
\begin{enumerate}
\item The SciERC dataset~\cite{luan-etal-2018-multi} annotates entities, relations, and coreference clusters in 500 abstracts from 12 AI conference and workshop proceedings indexed in the Semantic Scholar Corpus. It includes a total of $8,089$ distinct named entities.
\item The ScienceExam dataset~\cite{smith-etal-2020-scienceexamcer} comprises entities from science exam questions aimed at students from 3rd to 9th grade, totaling $133,000$ entities.
\item The ScienceIE dataset~\cite{augenstein2017semeval} provides $9,946$ entities appeared in 500 scientific paragraphs.
\end{enumerate}

For CLS tasks, we use the following 3 standard benchmark datasets.
\begin{enumerate}
\item The SciERC dataset~\cite{luan-etal-2018-multi} includes $4,716$ annotated sentences categorized for classification according to relations such as USED-FOR, FEATURE-OF, PART-OF, etc. 
\item The ACL-ARC dataset~\cite{jurgens2018measuring} comprises $1,941$ citation instances from 186 papers in the ACL Anthology Reference Corpus, with each instance annotated by domain experts with categories such as Future, Background, Motivation, etc. 
\item The SciCite dataset~\cite{cohan-etal-2019-structural} encompasses $11,020$ citation instances from $6,627$ scholarly papers, with each citation classified into categories such as Background, Method, and Result Comparison. 
\end{enumerate}
While the 3 benchmark datasets are popular, they all involve sentence-level classifications. We therefore formulate another 3 CLS tasks that use entire abstracts of arXiv papers as input. They are predictions of: 
\begin{enumerate}
\setcounter{enumi}{3}
\item major category of a paper (8 categories including physics, computer science, etc.); 
\item sub-category of a paper (172 labels including Astrophysics of Galaxies, Artificial Intelligence, etc.); 
\item whether a paper spans more than one major category, which can be considered as an interdisciplinary paper, as assessed by the submitting author. 
\end{enumerate}
The last task is motivated by the increasing prevalence of interdisciplinary science~\cite{gates2019nature} and its role in tackling important societal problems~\cite{ledford2015how}. 

Finally, for REL tasks, we use the abstracts of two papers to predict whether they are in the same broad category.

\subsection{Experimental setup}

We compare our AnnualBERT models with other scientific BERT models, including BioBERT~\cite{lee2020biobert}, SciBERT~\cite{beltagy2019scibert}, BioMedBERT~\cite{gu2021domain}, as well as the original BERT~\cite{devlin2018bert}. For our evaluations, we utilize the AnnualBERT model corresponding to the dataset's year of release. 

To ensure consistent comparisons across various BERT variants, we standardize the training configuration for each model. For the NER tasks, we adopt the NER model structure from \texttt{SimpleTransformers}, which incorporates a direct linear layer, and set the training parameters as follows: maximum input length of 512, training duration of 10 epochs, and batch size of 32. The best model parameters after each epoch are saved. For the CLS tasks, the vectors obtained using the \texttt{[CLS]} token pooling strategy are used as the input into a multilayer perceptron (MLP) for classification during the fine-tuning phase. The primary training parameters are set as follows: a training duration of 4 epochs, a batch size of 16, and weight decay at 0.01, and the best-performing model on the development set is saved every 100 steps. For the REL task, we use the dual tower structure~\cite{reimers2019sentence}. We feed the abstracts of two arXiv papers into the identical BERT-like model and employ the \texttt{[CLS]} tag pooling strategy to obtain their respective feature vectors, $u$ and $v$, that effectively represent each abstract. We then concatenate $u$ and $v$ and their element-wise absolute difference $\left\vert u-v \right\vert$. This concatenated vector is then fed into a MLP layer for classification. 

We conduct all the experiments on a workstation with two Nvidia A6000 GPUs. 

\subsection{Experimental results}

Table~\ref{tab:benchmark_results} presents the experimental results of various variants of BERT with two different methods of training our AnnualBERT: continual ($\mathcal{M}_t$) and ``one-time'' ($\overline{\mathcal{M}}$). Firstly, both variants of our AnnualBERT do not perform well on the NER tasks, and interestingly, neither do the other two models pretrained from scratch (SciBERT and BioMedBERT). This might be due to the tokenization strategy we use during the pretraining process. It may be also related to the fact that BioBERT, the best-performing model, is resulted from continual pretraining from the general Wikipedia and books corpus, which may have a better alignment with the benchmark datasets like the ScienceExam. Secondly, focusing still on the NER tasks, while both variants of AnnualBERT models have similar performances on the SciERC and ScienceExam datasets, the continual training version has a much better performance than the model from one-time training on the ScienceIE dataset. 

\begin{table}[!t]
\centering
\caption{Experimental results of NLP tasks using different (scientific) BERT variants. Reported are F1 scores.}
\label{tab:benchmark_results}
\resizebox{\textwidth}{!}{
\begin{tabular}{c|c|c|c|c|c|c|c|c}
\toprule
Task & Field & Dataset & BERT & BioBERT & SciBERT & BioMedBERT & AnnualBERT $\mathcal{M}_t$ & AnnualBERT $\overline{\mathcal{M}}$ \\
\midrule
\multirow{6}{*}{CLS} 
& CS & SciERC & 85.96 & 87.18 & 87.21 & 86.96 & \textbf{87.60} & 86.49\\
& CS & ACL-ARC & 76.22 & 75.97 & 80.32 & 79.53 & 79.88 & \textbf{81.80}\\
& Multi & SciCite & 85.20 & \textbf{86.63} & 86.33 & 86.33 & 85.77 & 85.09\\
& Multi & arXiv (major) & 88.83 & 88.96 & 90.69 & 89.95 & \textbf{91.41} & 91.12 \\
& Multi & arXiv (sub) & 44.48 & 45.31 & 57.03 & 51.60 & \textbf{58.24} & 57.99\\
& {Cross-field identification} & arXiv & 83.58 & 85.42 & 87.85 & 85.82 & \textbf{92.13} & 83.99\\ 
\midrule
\multirow{3}{*}{NER} 
& CS & SciERC & 59.69 & \textbf{63.61} & 62.55 & 61.84 & 60.96 & 60.80\\
& Multi & ScienceExam & 80.53 & \textbf{82.88} & 82.13 & 80.69 & 77.69 & 78.28\\
& Multi & ScienceIE & 33.56 & \textbf{34.86} & 33.97 & 34.40 & 33.95 & 30.93\\
\midrule
\multirow{1}{*}{REL} 
& Multi & arXiv & 86.79 & 87.64 & 89.18 & 87.44 & {89.52} & \textbf{90.43}\\
\bottomrule
\end{tabular}
}
\end{table}

For the 3 sentence-level CLS tasks, on the other hand, our continually pretrained AnnualBERT has the best performance on the SciERC dataset and comparable performances to the other specialized BERT models on the ACL-ARC and SciCite datasets. In general, the performance differences among these models in the three tasks are rather small. However, significant performance gaps become evident for the abstract-level, three domain-specific tasks, where our continually pretrained AnnualBERT $\mathcal{M}_t$ performs the best. Particularly, for the sub-category prediction task, which involves classifying papers into one of the 172 sub-fields based on their abstracts, our AnnualBERT $\mathcal{M}_t$ significantly leads the other models including the one-time trained $\overline{\mathcal{M}}$. Notably, SciBERT also has a comparable performance in this task, likely due to its training corpus containing a substantial number of computer science papers. Finally, AnnualBERT (continual) has a dominant performance in the cross-field paper identification task, indicating its superior knowledge representation ability for interdisciplinary papers that stretch multiple domains, while the other models, despite fine-tuned, struggle with this task. Interestingly, while the AnnualBERT (one-time) shows relatively similar performances compared with its continual variant on most datasets, it performs significantly worse than the continual version and even other models on the task of cross-field paper identification. This disparity suggests that organizing the corpus by year, as done in the continual variant, may enhance the model's ability to capture interdisciplinary knowledge. By incrementally exposing the model to cross-domain data over time, rather than feeding it all at once, our yearly-based continual training strategy appears to better equip language models for tasks spanning multiple domains. Taken together, these results substantiate the validity of our AnnualBERT models and the effectiveness of domain-specific pretraining for domain-specific tasks, in consistence with prior studies~\cite{gu2021domain}.

\section{Citation link prediction}

Another perspective to assessing capabilities of our AnnualBERT models in learning and encoding semantic information of papers is through link prediction tasks in citation network. The rationale behind using this task lies in the nature of scientific discourse, where citations are formed by not only topical alignment but also nuanced interplay of numerous other factors like ideas and methodologies that are shared between two papers. Therefore, the ability of a language model in encoding abstracts into feature vectors that are highly predictive of whether two papers are related through citations reflects the model's proficiency in grasping those complex semantic relationships, which in turn are instrumental for generating meaningful insights into domain-specific research~\cite{peng2021neural}. 

As our setup requires a citation network between arXiv papers, we derive such a network from the Microsoft Academic Graph (MAG) dataset~\cite{sinha2015overview}. Doing so can avoid the otherwise extensive efforts of matching each referenced entry within the arXiv corpus, which may involve identifying the published version of a arXiv paper. To build our citation network, we first identify the corresponding MAG version of each arXiv paper, by extracting arXiv IDs from paper URLs provided by MAG, and then use the citation relationships between MAG papers to construct the network. Through this method, we identify nearly 1.5 million arXiv papers and 14.8 million citation relationships between them. 

Given the constructed citation network, we study link prediction in both static and temporal settings.  

\subsection{Link prediction in static network}

Let $G = (V, E)$ denote the static citation network observed at 2020, where $V$ is the set of $N = \left\vert V \right\vert$ nodes and $E = \{(u, v)\}$ is the set of links indicating citation relationships between two arXiv papers $u, v \in V$. We study link prediction under the supervised learning setting. That is, we learn from data a model mapping from node-pair features to link missingness. Here we train classifiers using 5-fold cross-validation by forming datasets $\mathcal{D} = (E^+, E^-)$ that are composed of positive training examples as a subset of observed links $E^+ \subset E$ and negative examples generated from sampling the same number of non-existent links $E^- \subset V \times V - E$. 

\subsubsection{Methods}

Link prediction in networks has been extensively studied in the past~\cite{liben2007link, ghasemian2020stacking}. Here we consider two families of approaches to predicting missing links: topological predictors and graph neural networks (GNN). 

\paragraph{Topological predictors}

Classical methods consider structural information in a network and assign a score $s(u, v)$ for a pair of nodes $u$ and $v$. We use the following 6 topological predictors.
\begin{enumerate}
\item Common neighbors is the number of shared neighbors between $u$ and $v$, \emph{i.e.}, $s = \left\vert \Gamma(u) \cap \Gamma(v) \right\vert$, where $\Gamma(\cdot)$ denotes the neighbors of a node. 
\item Jaccard coefficient (JC) is the normalized overlap between the neighbors: $s = \frac{\left\vert \Gamma(u) \cap \Gamma(v) \right\vert}{\left\vert \Gamma(u) \cup \Gamma(v) \right\vert}$. 
\item Preferential attachment (PA) index captures the tendency that more connected nodes are more likely to attract new links: $s = \left\vert \Gamma(u) \right\vert \times \left\vert \Gamma(v) \right\vert$.
\item The Adamic/Adar (AA) index~\cite{liben2007link} defines the score as $s = \sum_{z \in \Gamma(u) \cap \Gamma(v)} \frac{1}{\log \left\vert \Gamma(z) \right\vert}$. It considers the importance of a common neighbor $z$, by giving more weight if $z$ has fewer connections, capturing the intuition that common neighbors in a less-populated field can be more significant in predicting a link between $u$ and $v$. 
\item The resource allocation (RA) index~\cite{zhou2009predicting} defines the score as $s = \sum_{z \in \Gamma(u) \cap \Gamma(v)} \frac{1}{\left\vert \Gamma(z) \right\vert}$. It measures the likelihood of a link between two nodes based on the principle of resource allocation, where nodes share resources through their common neighbors. 
\item Personalized PageRank (PPR) corresponds to the $v$-th entry in the stationary distribution of a random walk with restart from $u$. Compared with the above predictors, PPR utilizes more global structural information in a network. 
\end{enumerate}
For all these methods, we use logistic regression. 

\paragraph{GNN}

One limitation of topological predictors is that they are unable to exploit node features, which in our context are the abstract of a paper. The rapid development of GNN has made it possible to integrate graph topology and node features for downstream tasks like link prediction. Here we leverage GraphSAGE~\cite{Hamilton2017InductiveRL} to generate embeddings $H^{(L)}$ of the nodes in our arXiv citation network, which are then used for link prediction. Note that our purpose is to evaluate different scientific BERT models rather than GNNs, and we choose GraphSAGE due to its popularity. Different from earlier graph embedding methods that are transductive, GraphSAGE is an inductive graph representation learning algorithm and learns aggregation functions so that it can inductively generate embeddings of unseen nodes. 

The same as other GNNs, GraphSAGE generates embeddings of a node by sampling and aggregating embeddings of its local neighbors. In our experiment, given an initial node embedding matrix $H^{(0)}$, we use graph convolutional network (GCN) as the aggregator: 
\begin{equation}
H^{(l)} = \sigma \left( \widetilde{D}^{-\frac{1}{2}} \widetilde{A} \widetilde{D}^{-\frac{1}{2}} H^{(l-1)} W^{(l-1)} \right) \, ,
\end{equation}
where $H^{(l)}$ represents the updated node representations after the $l$-th layer of GraphSAGE, $\sigma$ is the activation function, $\widetilde{D}$ is the degree matrix of the network $G$, $\widetilde{A}$ is the adjacency matrix of the network with added self-loops, and $W^{(l-1)}$ is a trainable weight matrix at layer $l-1$. We set the number of layers $L$ to 2 and use 2 MLP layers for modeling interactions between embeddings of node $u$ and $v$: 
\begin{equation}
\begin{split}
s(u, v) &= MLP \left( h_{u}^{(L)}, h_{v}^{(L)} \right) \\
        &= w_{2} \sigma \left( w_{1} \left[ h_{u}^{(L)}; h_{v}^{(L)} \right] + b_{1} \right) + b_{2} \, .
\end{split}
\end{equation}
The loss function is the binary cross-entropy loss:
\begin{equation}
\mathcal{L} = - \sum_{u, v \in V} \left( 
    y_{u \sim v} \log(s(u, v)) + 
    (1 - y_{u \sim v}) \log(1 - s(u, v)) 
\right) \, ,
\end{equation}
where $y_{u \sim v}$ is 1 if there is a link between $u$ and $v$, and 0 otherwise. 

We consider the following 9 ways to obtain $H^{(0)}$. 
\begin{enumerate}
\item We associate each node with a random $768$-dimensional feature vector, thus $H^{(0)} \in \mathbb{R}^{N \times 768}$. The purpose is to, by feeding no useful node features to GraphSAGE, make it learn only topological information, so that we can compare with the above topological predictors. 
\item We generate the one-hot encoding matrix for major category of papers as the $H^{(0)} \in \mathbb{R}^{N \times 8}$ (8 categories). This is based on the fact that citations tend to occur within the same field, and injecting field label information to GraphSAGE would improve its citation link prediction performance, thereby serving as a strong baseline to check to what extent paper abstract provides additional information than paper category does. 
\item Similarly, we use the one-hot encoding matrix for paper sub-category labels as the $H^{(0)} \in \mathbb{R}^{N \times 172}$ (172 subcategories). 
\item The rest 6 feature matrices $H^{(0)} \in \mathbb{R}^{N \times 768}$ are encoded by each of the 6 scientific BERT models using the \texttt{[CLS]} tag pooling strategy on paper abstracts. For our models, we use AnnualBERT $\mathcal{M}_{2020}$ and $\overline{\mathcal{M}}$. By using different feature matrices encoded by the surveyed models, we can ascertain whether our AnnualBERT is the most useful encoder in predicting citation relationship between papers. 
\end{enumerate}

\subsubsection{Experimental results}

Table~\ref{tab:link-pred-static} reports the experimental results, providing a comprehensive comparison of various link prediction methods applied to our arXiv citation network. Firstly, the results indicate that some topological predictors---AA and RA---obtain the highest (possible) precision, highlighting their effectiveness in predicting connections between academic papers. However, these methods have rather low recall, suggesting that they miss many true positives. PPR appears to balance the best between precision and recall, considering especially its much higher recall than the other topological predictors. 

\begin{table*}[!t]
\centering
\caption{Experimental results for link prediction in the static arXiv citation network.}
\label{tab:link-pred-static}
\resizebox{\textwidth}{!}{
\begin{tabular}{lccccc} 
\toprule
Model & AUC-ROC & Accuracy & Precision & Recall & F1 Score \\ 
\midrule
Common Neighbors        & 0.739 (0.000) & 0.670 (0.000) & 0.949 (0.000) & 0.266 (0.000) & 0.506 (0.000) \\
Jaccard Coefficient     & 0.739 (0.000) & 0.739 (0.000) & 0.997 (0.000) & 0.333 (0.001) & 0.648 (0.000) \\
Adamic/Adar             & 0.708 (0.000) & 0.708 (0.000) & \textbf{1.000} (0.000) & 0.148 (0.001) & 0.589 (0.000) \\
Preferential Attachment & 0.895 (0.000) & 0.820 (0.000) & 0.978 (0.000) & 0.235 (0.000) & 0.813 (0.000) \\
Resource Allocation     & 0.740 (0.000) & 0.738 (0.000) & \textbf{1.000} (0.000) & 0.266 (0.007) & 0.646 (0.000) \\
Personalized PageRank   & 0.870 (0.000) &0.776 (0.000)&0.814 (0.015)&0.651 (0.033) & 0.767 (0.000)\\ 
\midrule
GraphSAGE+Random                              & {0.827 (0.013)} & {0.749 (0.013)} & {0.772 (0.012)} & {0.706 (0.021)} & {0.737 (0.016)}  \\
GraphSAGE+Major category                      & {0.858 (0.014)} & {0.782 (0.003)} & {0.722 (0.002)} & {0.914 (0.004)} & {0.807 (0.003)}  \\
GraphSAGE+Subcategory                         & {0.975 (0.002)} & {0.928 (0.003)} & {0.940 (0.003)} & {0.917  (0.002)} & {0.928 (0.003)}  \\
GraphSAGE+BERT                                & {0.984 (0.002)} & {0.936 (0.003)} & {0.954 (0.006)} & {0.915 (0.002)} & {0.934 (0.002)}  \\
GraphSAGE+BioBERT                             & 0.978 (0.001) & 0.923 (0.002) & 0.941 (0.003) & 0.902 (0.003) & 0.921 (0.002) \\
GraphSAGE+SciBERT                             & 0.984 (0.002) & 0.936 (0.003) & 0.954 (0.006) & 0.915 (0.002) & 0.934 (0.002) \\
GraphSAGE+BioMedBERT                          & 0.980 (0.005) & 0.928 (0.008) & 0.945 (0.010) & 0.910 (0.006) & 0.927 (0.008) \\
GraphSAGE+AnnualBERT $\mathcal{M}_{2020}$     & \textbf{0.991} (0.002) & \textbf{0.955} (0.004) & 0.967 (0.006) & \textbf{0.942} (0.003) & \textbf{0.955} (0.004) \\
GraphSAGE+AnnualBERT $\overline{\mathcal{M}}$ & 0.983 (0.002) & 0.933 (0.003) & 0.951 (0.005) & 0.912 (0.003) & 0.932 (0.003) \\
\bottomrule
\end{tabular}
}
\end{table*}

Secondly, turning to GraphSAGE-based methods with different initial node feature matrices $H^{(0)}$, we find that feeding no useful node features (random $H^{(0)}$) results in performance similar to PPR. Providing paper major category information to GraphSAGE significantly improves recall (0.914), while overall performance metrics (AUC-ROC and F1) remain largely similar to PPR. This meets with our expectation that major category is a highly useful discriminative predictor in distinguishing pairs of papers with or without citations. Supplementing sub-category information to GraphSAGE further improves performance in a great extent, with the AUC-ROC increasing from 0.858 to 0.975. Such a improvement is mainly attributed to the boost in precision from 0.722 to 0.940, while the recall stays almost unchanged around 0.91. These results make this simple representation method a highly competitive alternative to sophisticated language models. 

Thirdly, using all the 5 different BERT models further improves AUC-ROC, implying that features encoded by them indeed provide additional information than subcategory does. Our AnnualBERT model $\mathcal{M}_{2020}$ stands out distinctly, achieving the highest score in AUC-ROC (0.991). Again, the same with the subcategory case, the increases in AUC-ROC stem from higher precision for all the BERT models except our AnnualBERT, as we can see that their recall remains constant around 0.91. Our AnnualBERT is the only model that improves both precision (0.967) and recall (0.942). 

Fourthly, comparing one-time trained $\overline{\mathcal{M}}$ with continually trained $\mathcal{M}_{2020}$ reveals two findings: (1) The performance of $\overline{\mathcal{M}}$---despite pretrained on the in-domain corpus---is comparable with other models pretrained on out-of-domain corpora; and (2) our proposed continual pretraining based on temporally organized corpora is a more effective strategy, with the resulting model $\mathcal{M}_{2020}$ performing better than the one-time trained model $\overline{\mathcal{M}}$, even though they use identical pretraining corpus. In summary, these results demonstrate the effectiveness of our model in learning and encoding semantic information of papers among all the evaluated methods. 

\subsection{Link prediction in temporal network}

\subsubsection{Experimental setup}

The arXiv citation network is actually a temporal (growing) network, as we know the publication year of each paper. This raises the question of how different models perform in a more challenging setting of prospective link prediction, where we train a classifier using links and non-links observed during an earlier period and then apply it to predict the appearance of future links. Formally, let $G_{t_0} = (V_{t_0}, E_{t_0})$ denote the citation network observed at year $t_0$, where the node set $V_{t_0} = \{ v \vert t_v \leq t_0 \}$ represents all the arXiv papers published until $t_0$ and $E_{t_0} = \{ (u, v) \vert u, v \in V_{t_0} \}$ is the set of citation links between them. With $G_{t_0}$, we use GraphSAGE and MLP to model interactions between embeddings of two arXiv papers in $V_{t_0}$, following the same process as in the static network case. We then apply the resulting architecture to predict links appeared in year $t_0 + \Delta t$, where $\Delta t$ denotes the number of years into the future. Specifically, we denote the test dataset as $\mathcal{D}_\text{test} = (B_{t_0+\Delta t}^+, B_{t_0+\Delta t}^-)$, where $B_{t_0+\Delta t}^+$ contains all the observed citations (positive examples) from papers published in year $t_0 + \Delta t$ to papers published until $t_0 + \Delta t$, \emph{i.e.}, $B_{t_0+\Delta t}^+ = \{ (u, v) \vert t_u = t_0+\Delta t, t_v < t_0 + \Delta t \}$, and $B_{t_0+\Delta t}^-$ is the set of non-existent links (negative examples) that are generated by, for each observed link $(u, v) \in B_{t_0+\Delta t}^+$, randomly choosing a different endpoint $v'$ published until $t_0$, \emph{i.e.}, $B_{t_0+\Delta t}^- = \{ (u, v') \vert t_u = t_0+\Delta t, t_{v'} < t_0, (u, v') \not \in B_{t_0+\Delta t}^+ \}$. 

In our experiment, we consider $t_0 = 2014$ and vary $\Delta t$ from 1 to 6, thus predicting future links appeared from 2015 to 2020. The network $G_{t_0}$ has $\left\vert V_{t_0} \right\vert = 637,740$ nodes and $\left\vert E_{t_0} \right\vert = 3,999,781$ links. We evaluate the performances of the above mentioned 8 ways of obtaining initial node embeddings $H^{(0)}$ by testing on the $\mathcal{D}_\text{test}$ datasets. For our method, we take AnnualBERT $\mathcal{M}_{t_0}$ for feature encoding. Here we do not test the one-time trained model $\overline{\mathcal{M}}$, since it contains future information. 

\subsubsection{Experimental results}

\begin{figure}[t!]
\centering
\includegraphics[trim=0 5mm 0 0, width=\linewidth]{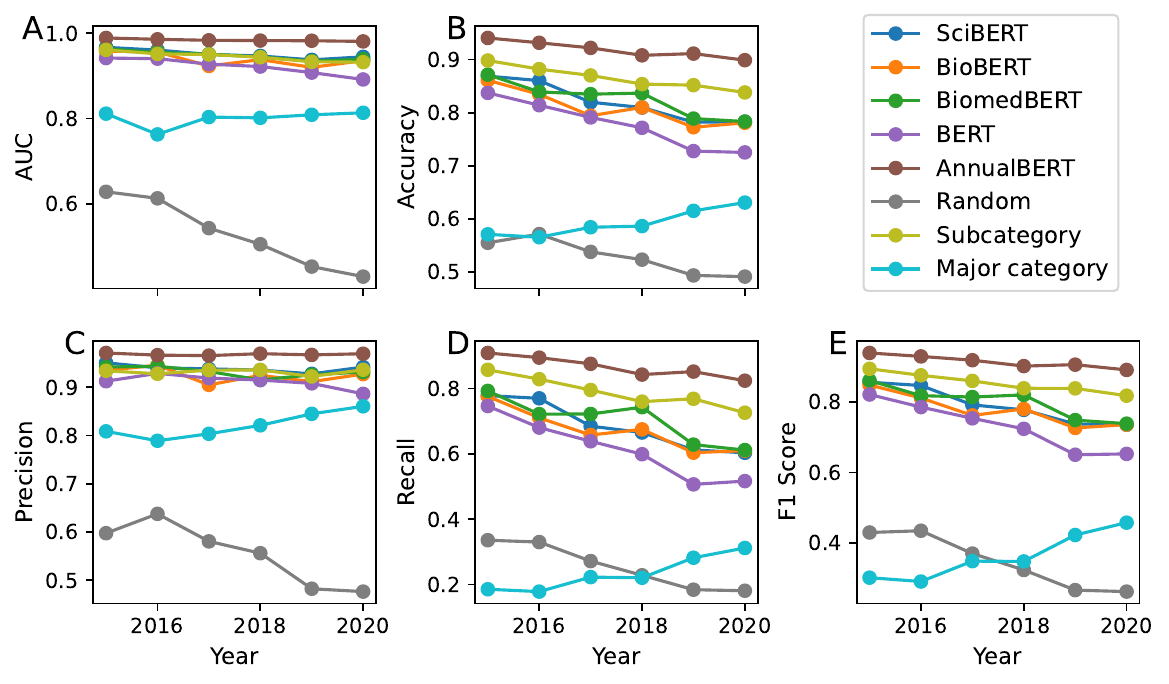}
\caption{Experimental results for link prediction in the temporal arXiv citation network.}
\label{fig:lp-temporal}
\end{figure}

Fig.~\ref{fig:lp-temporal} presents the experimental results, comparing the performances of the 8 different methods for generating the $H^{(0)}$ representations. Overall, the performance of most methods declines as the time shift $\Delta t$ increases, aligning with the expectation of a growing challenge of predicting future links over a longer period. However, the performance decay is largely driven by the deterioration of recall (Fig.~\ref{fig:lp-temporal}D), while the precision remains stable (Fig.~\ref{fig:lp-temporal}C). This highlights the difficulty in retrieving pairs of papers with citation relationships (true positives), potentially pointing to the emergence of scientific innovation resulted from making novel combinations~\cite{uzzi2013atypical, ke2020technological}. This could serve as a future direction to explore the relationship between predictability of citation pairs and scientific impact. Looking at different methods, the two encoders with limited information (\emph{i.e.}, random and major category) perform poorly across all metrics, as they lack the contextual and structural information necessary for effective link prediction. By contrast, the subcategory-based method yields surprisingly strong results, even surpassing most methods except AnnualBERT in terms of recall and F1. This impressive performance may be attributed to the specificity of the subcategories, with a total of 172 distinct subcategories providing highly targeted information. Our AnnualBERT model still consistently outperforms with a significantly large gap when comparing with the other BERT-based models, achieving the highest scores in AUC-ROC, accuracy, recall, and F1 Score, underscoring the effectiveness of domain-specific pretraining in capturing richer semantic and temporal representations, making it particularly robust for long-term link prediction. 

\section{Mining AnnualBERT} \label{sec:series} 

In the previous two sections, we have demonstrated the validity of our AnnualBERT series of models ${\mathcal{M}_t}$ and their superior performances than the other scientific BERT models for tasks specific to the arXiv domain. In this section, we are interested in mining these models, as we reason that each of the models serves as a condensed representation of the scientific literature over a specific period, the mining of which would enable us to understand the changing landscape of science as captured in the arXiv corpus. 

\subsection{Temporal adaption} \label{subsec:temp-adapt}

\subsubsection{Motivation}

\begin{figure}[t!]
\centering
\includegraphics[trim=0 5mm 0 0, width=1\linewidth]{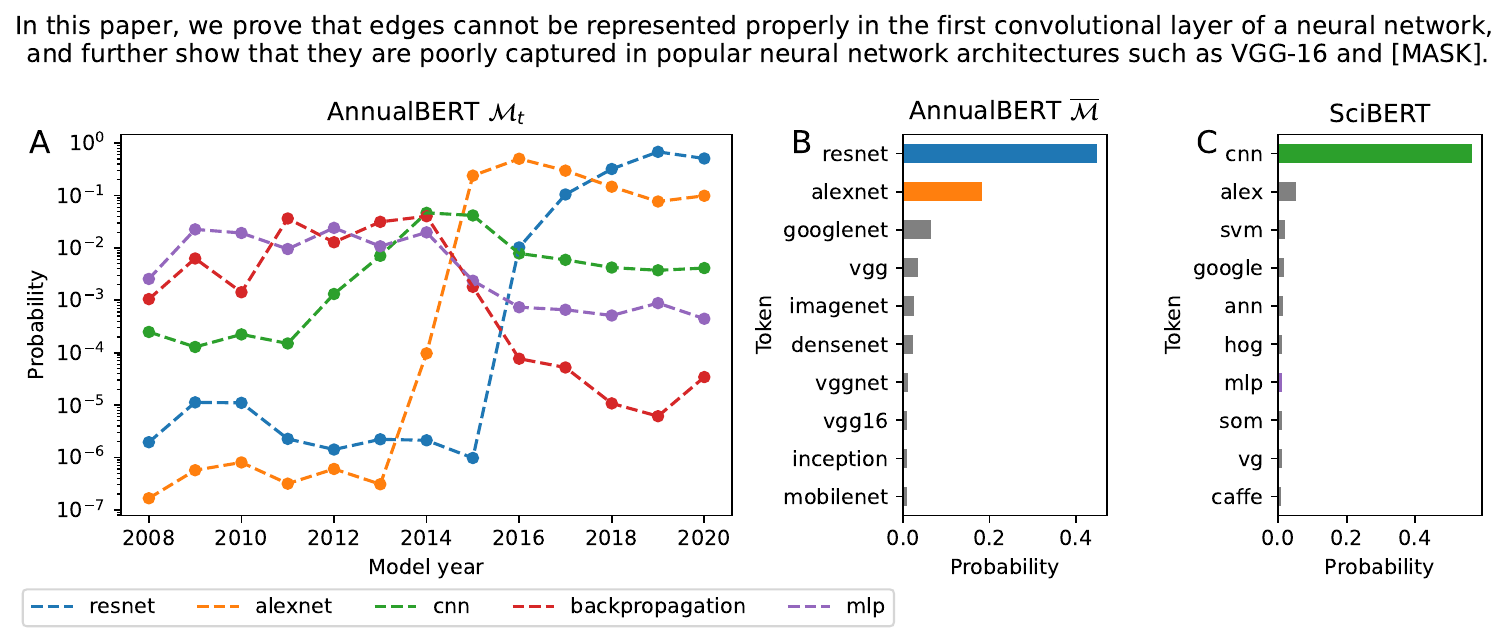}
\caption{The probabilities of top tokens given by different models, for the sentence shown at the top.}
\label{fig:word_prob}
\end{figure}

As mentioned above, our choice of using whole word rather than subword as token is motivated by the fact that words are the most basic constituent unit for ideas, and the probabilistic feature of BERT to predict the masked token given its context (other tokens in the sentence) allows us to analyze the dynamics of research by examining how the probability of each word change over time, as assessed by our AnnualBERT model series. For example, consider the sentence shown in Fig.~\ref{fig:word_prob} about edge representation in image models, we mask ``ResNet'' and ask each of the AnnualBERT models $\mathcal{M}_t$ to output the most likely tokens and their associated probabilities (Fig.~\ref{fig:word_prob}A). For comparisons, we also test with the one-time trained $\overline{\mathcal{M}}$ (Fig.~\ref{fig:word_prob}B) and SciBERT (Fig.~\ref{fig:word_prob}C). For SciBERT, as it uses subword tokenization, meaningful ideas like ``ResNet'', ``AlexNet'', ``GoogLeNet'', and ``VGG'' are all broken into subwords like ``alex'', ``google'', and ``vg''. Consequently, SciBERT is not able to make the correct guess, leaving ``cnn'' as the most likely token and ``sub-ideas'' like ``alex'' and ``google'' small probabilities. By contrast, $\overline{\mathcal{M}}$ adopts whole-word tokenization, and most of the top choices are image models. Yet, the probability of ``ResNet'' (0.45) does not overwhelmingly dominate that of ``AlexNet'' (0.18), indicating that the latter one is still a viable choice. Finally, annually trained models $\mathcal{M}_t$ consider further the temporal dimension and reveal the dynamics of the competition between AlexNet and ResNet. In summary, this example highlights the rationale and the advantage of whole-word tokenization for the analysis of scientific texts where new terminologies are constantly emerging. 

\begin{figure*}[t!]
\centering
\includegraphics[trim=0 5mm 0 0, width=\linewidth]{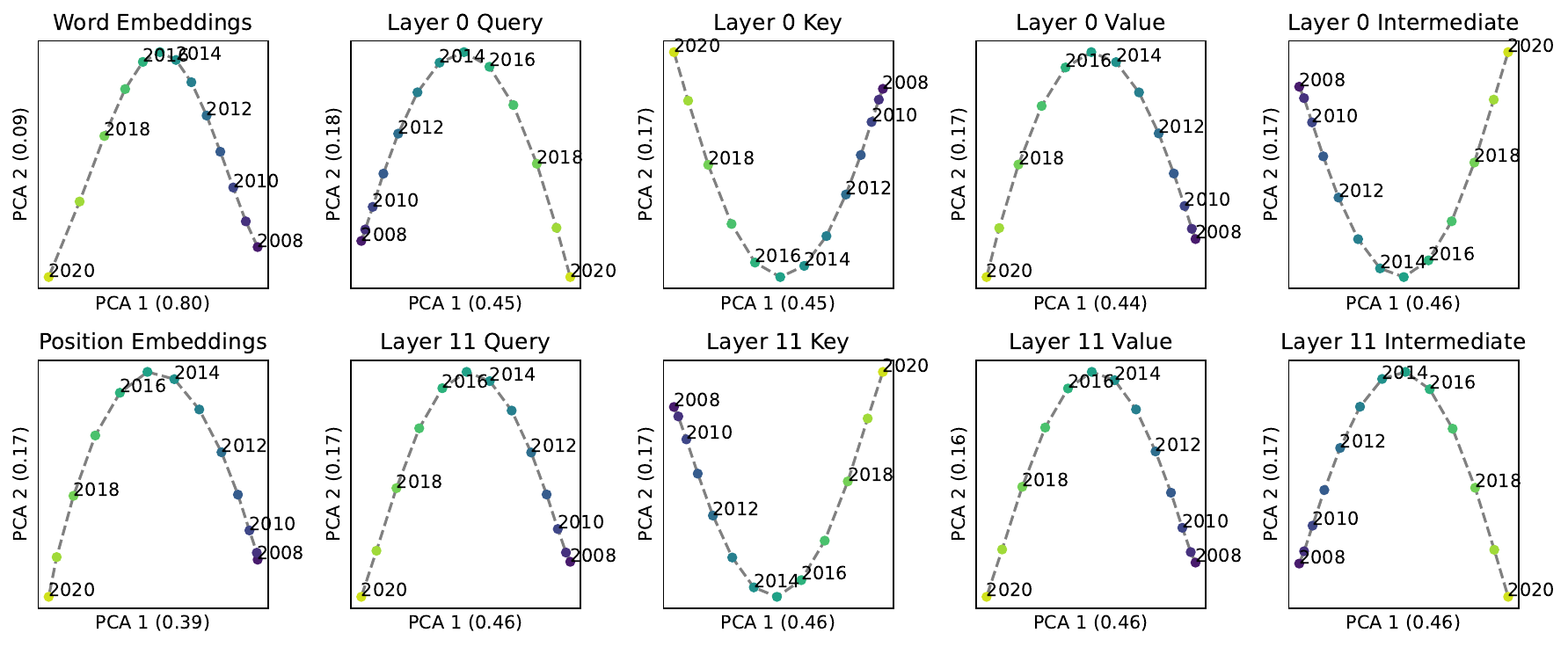}
\caption{PCA visualizations of model weights in different layers.}
\label{fig:pca_model_weight}
\end{figure*}

This analysis raises a question: How do the model's internal weights evolve to capture such contextual shifts in word usage over time? To explore this, we examine how the model weights change as AnnualBERT adapts to new corpora published annually. Specifically, we focus on visualizing the weights of different layers in the model. To this end, we select various layers to visualize their weights. For instance, in the word embedding layer, we transform each model's \(53100 \times 768\) matrix associated with that layer into a one-dimensional vector, resulting in a \(13 \times 40780800\) matrix. Subsequently, we employ principal component analysis (PCA) to reduce the dimensionality of the flattened word embedding weights to two dimensions. The same procedure is applied to other layers to obtain visualizations shown in Fig.~\ref{fig:pca_model_weight}, from which we draw two observations. 
\begin{enumerate}
\item \textbf{Redundancy in model weights}: For most layers, the first two principal components explain over 60\% of the variance, with the word embedding layer exhibiting even higher redundancy (nearly 90\%). This suggests that the model weights are highly compressible, aligning with findings on the distillability of large language models~\cite{liu2022learning}. 
\item \textbf{Temporal trajectory of weights:} The weight space across layers exhibits a distinct quadratic pattern, with model weights from different years lying along a smooth curve. This indicates that the model parameters undergo systematic changes as they adapt to new corpora, implying the temporal evolution of scientific language.
\end{enumerate}
These results support the hypothesis that continual pretraining on corpora formed by time leads to constant adaptations in the model's knowledge representation. This finding is important for understanding how changes in scientific literature are encapsulated within the evolving parameters of domain-specific language models like AnnualBERT. 

\subsubsection{Experimental setup}

In continual learning, learning new knowledge often leads to forgetting old knowledge, which is manifested as improved performance on new tasks but decreased performance on old ones. Different from the previous focus, we concentrate on temporal differences. Particularly, the observation that model weights change when adapting to timestamped corpora within the same domain naturally leads to the questions of how both learning and forgetting manifest and how we can quantify the extent of knowledge forgetting. We answer them in the context of prediction tasks. 

Given a specific prediction task, for each data year $\tau$, we randomly sample $1,600$ and 200 abstracts published in $\tau$ respectively as the training and test set, extract their feature vectors from the \texttt{[CLS]} tokens generated by each model $\mathcal{M}_t$, and train a Random Forests model using the training set and test it on the test set, resulting a performance matrix $P_{t,\tau}^{(r)}$. To mitigate the effects due to random sampling, we repeat this process for 50 runs, using a different random seed for each run $r$. We then summarize the performance matrices from individual runs into one summary matrix, by (1) averaging the matrices $\overline{P}_{t,\tau} = \sum_{r} P_{t,\tau}^{(r)} / 50$; (2) performing column wise (data year $\tau$) min-max normalization of $\overline{P}_{t,\tau}$ into $\widetilde{P}_{t,\tau}$, to account for the inherent variability in classification difficulty of individual data years; and (3) performing column wise subtraction of the diagonal elements in $\widetilde{P}_{t,\tau}$ from $\widetilde{P}_{t,\tau}$, to obtain relative performances compared to the $t = \tau$ cases. We denote the final matrix as $\widehat{P}_{t,\tau}$. 

Below we consider three tasks: predicting major and sub-category of a paper and whether it is interdisciplinary, and the raw performance measure is F1-score. 

\subsubsection{Experimental results}

\begin{figure*}[t!]
\centering
\begin{minipage}[b]{0.5\textwidth}
    \centering
    \includegraphics[trim=0 10mm 0 0, width=0.87\linewidth]{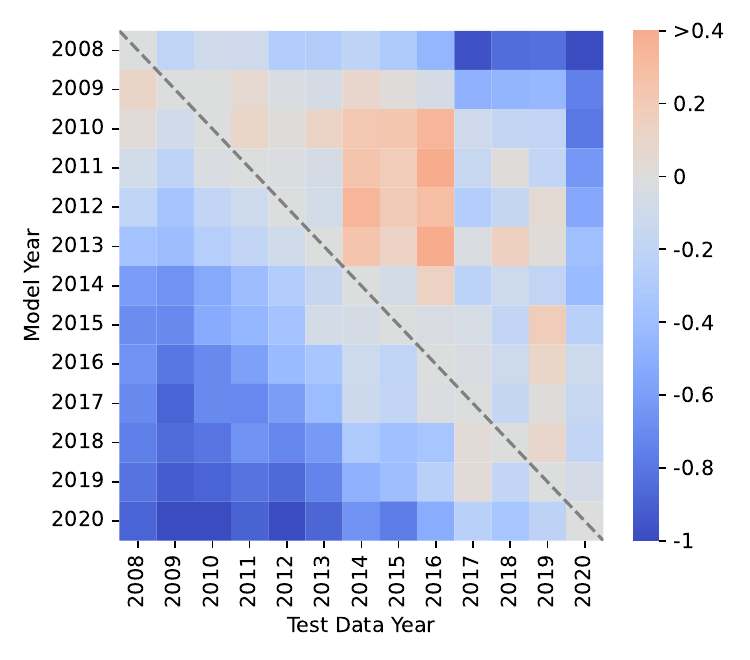}
    \caption*{(a)}
    \label{fig:heatmap_major}
\end{minipage}%
\hfill
\begin{minipage}[b]{0.5\textwidth}
    \centering
    \includegraphics[trim=0 10mm 0 0, width=0.87\linewidth]{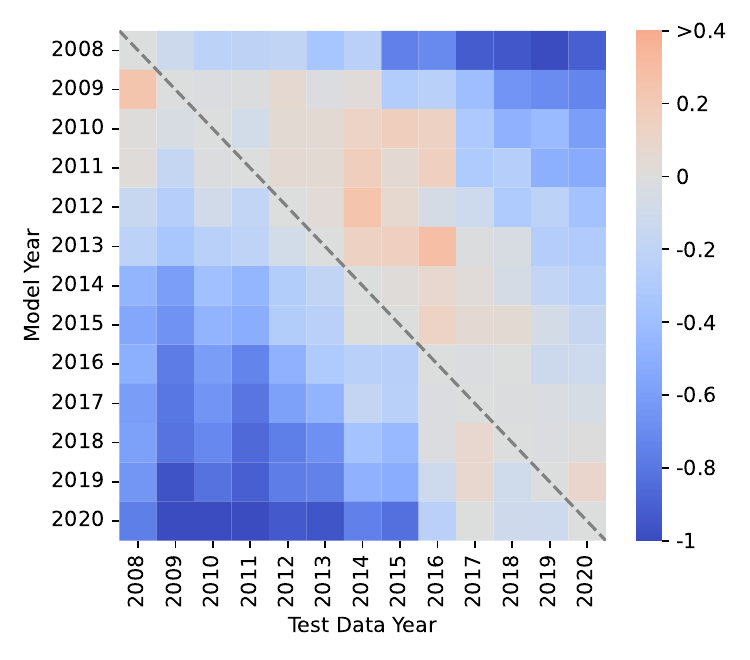}
    \caption*{(b)}
    \label{fig:heatmap_major_cut}
\end{minipage}

\begin{minipage}[b]{0.5\textwidth}
    \centering
    \includegraphics[trim=0 10mm 0 0, width=0.85\linewidth]{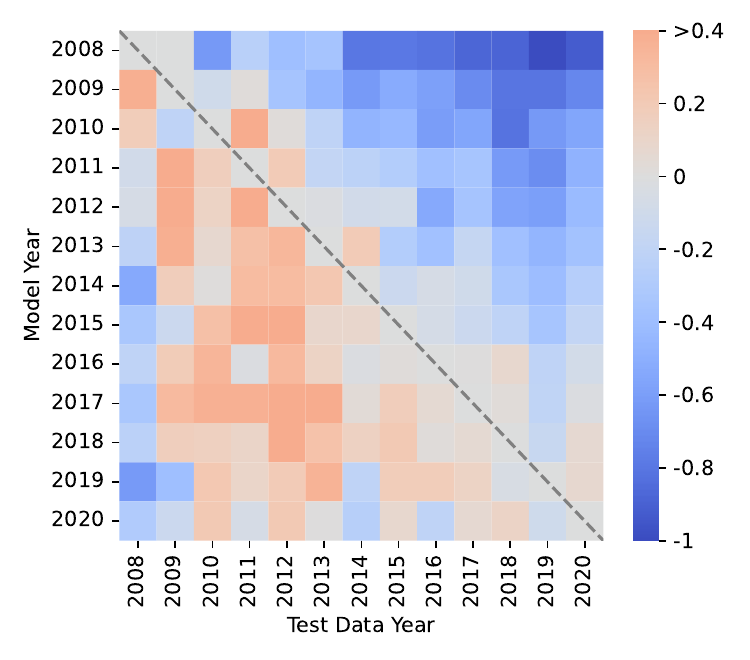}
    \caption*{(c)}
    \label{fig:heatmap_minor}
\end{minipage}%
\hfill
\begin{minipage}[b]{0.5\textwidth}
    \centering
    \includegraphics[trim=0 10mm 0 0, width=0.85\linewidth]{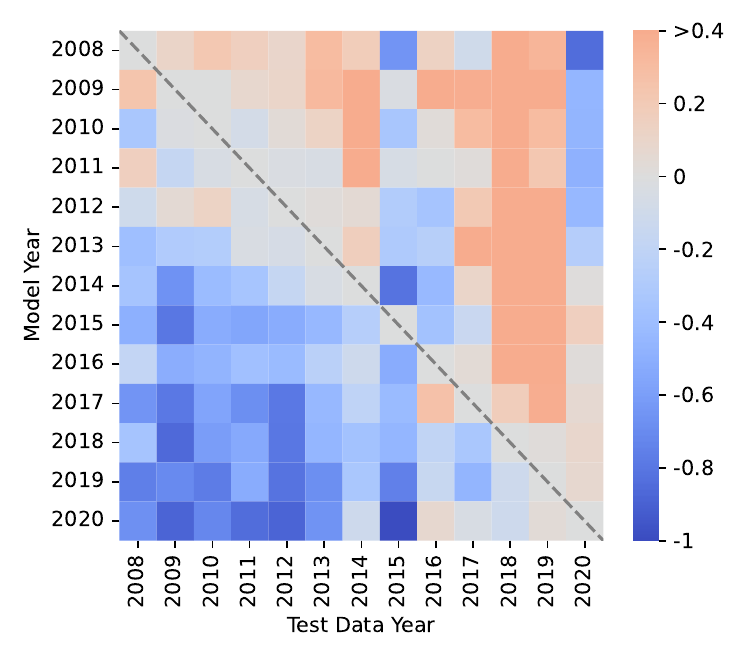}
    \caption*{(d)}
    \label{fig:heatmap_cross}
\end{minipage}
\caption{Summary F1 matrices $\widehat{P}_{t,\tau}$ showing how AnnualBERT models $\mathcal{M}_t$ perform in prediction tasks on papers from different data years $\tau$. (a--b) Major category prediction based on (a) entire abstracts and (b) only second half of abstracts. (c) Subcategory classification; (d) Cross-field identification.}
\label{fig:heatmaps_classification}
\end{figure*}

Fig.~\ref{fig:heatmaps_classification}a presents $\widehat{P}_{t,\tau}$ for the major category prediction task. The lower triangular part corresponds to cases where a model $\mathcal{M}_t$ predicts the major category of a ``past'' abstract ($t > \tau$). We observe under-performance, indicating that AnnualBERT does forget: As the model continues adapting to a newly published corpus, it captures semantic representation changes specific to that period while forgetting some knowledge from the past, such that the performances in the knowledge representation on the previous datasets decrease. This representation forgetting becomes more pronounced when the models make predictions of more distant past papers (darker blues moving away from the diagonal dashed line). 

The upper triangular part in Fig.~\ref{fig:heatmaps_classification}a corresponds to cases where a model $\mathcal{M}_t$ makes prediction on ``future'' abstracts ($t < \tau$). It demonstrates declining performances as well, due to the models' inability to encode an abstract not yet published. However, we notice that for the 2014--2016 test sets, the $\mathcal{M}_{2010}$--$\mathcal{M}_{2013}$ models perform better than the others. We hypothesize that this is because the cumulative nature of science advances is particularly evident, to the extent that there is a significant amount of flow of scientific discourse from full-text of past papers to background information in abstracts of future papers. To test this hypothesis, we re-run the entire prediction task, but using ``ablated'' training and test sets where we retain only the second half of the sentences in an abstract, through which we effectively, albeit roughly, reduce the impact of the scientific discourse flow. The results, which can be found in Fig.~\ref{fig:heatmaps_classification}b, show that there is a notable decline in the predictive performances for those years (the lighter red block), partially validating our hypothesis. 

Fig.~\ref{fig:heatmaps_classification}c presents $\widehat{P}_{t,\tau}$ for the subcategory prediction task. Similar to the previous task, predicting subcategory of future abstracts remains as a challenging task. However, different from Fig.~\ref{fig:heatmaps_classification}a, the phenomenon of representation forgetting for subcategory classification is not observed. On the contrary, there are noticeable performance increases when the models predict past abstracts (the lower triangular part). Such over-performance is not due to a smaller number of subcategory labels in earlier years, as the observation is actually reinforced if we focus on subcategories with the most papers. Instead we speculate that this might be attributed to the need for constant training corpora in order for the model to establish robust representation for specialized, fine-grained knowledge, and as the model progressively sees more relevant training data from subsequent years, its performance for these recent years improves. 

Fig.~\ref{fig:heatmaps_classification}d shows $\widehat{P}_{t,\tau}$ for the task of identifying cross-field papers. We note that the results are different from the previous two tasks. AnnualBERT under-performs in identifying past cross-field papers, indicating representation forgetting. Yet, it over-performs for future papers, suggesting that interdisciplinary papers could be published ahead of time and potentially shape the development of later papers. 

In summary, our results above paint a more complex picture about the representation forgetting phenomenon studied previously. We find that it is highly task dependent: While for tasks like major category prediction, AnnualBERT does forget, but for a slightly different task of predicting subcategory, representation forgetting is no longer evident. 

\subsection{Interpolation of models}

While we have found that AnnualBERT adapts to new corpora, Fig.~\ref{fig:pca_model_weight} in the meantime suggests that model weights are located close in the weight space, raising the question of whether we can interpolate models so that the interpolated ones $\mathcal{M}_{t_0}^{\Delta t}$ have similar performances with the ``real'' model $\mathcal{M}_{t_0}$ in prediction tasks. We derive the  interpolated model $\mathcal{M}_{t_0}^{\Delta t}$ ($\Delta t \geq 1$) by ``averaging'' $\mathcal{M}_{t_0 - \Delta t}$ and $\mathcal{M}_{t_0 + \Delta t}$, treating it as a linear system: 
\begin{equation}
\mathcal{M}_{t_0}^{\Delta t} = \frac{1}{2} \left( \mathcal{M}_{t_0 - \Delta t} \oplus \mathcal{M}_{t_0 + \Delta t} \right) \, .
\end{equation}
We then compare performances for the 3 prediction tasks with same settings as described before. 

\begin{figure*}[!t]
\centering
\begin{minipage}[b]{0.33\textwidth}
\centering
\includegraphics[trim=0 15mm 0 0, width=1\linewidth]{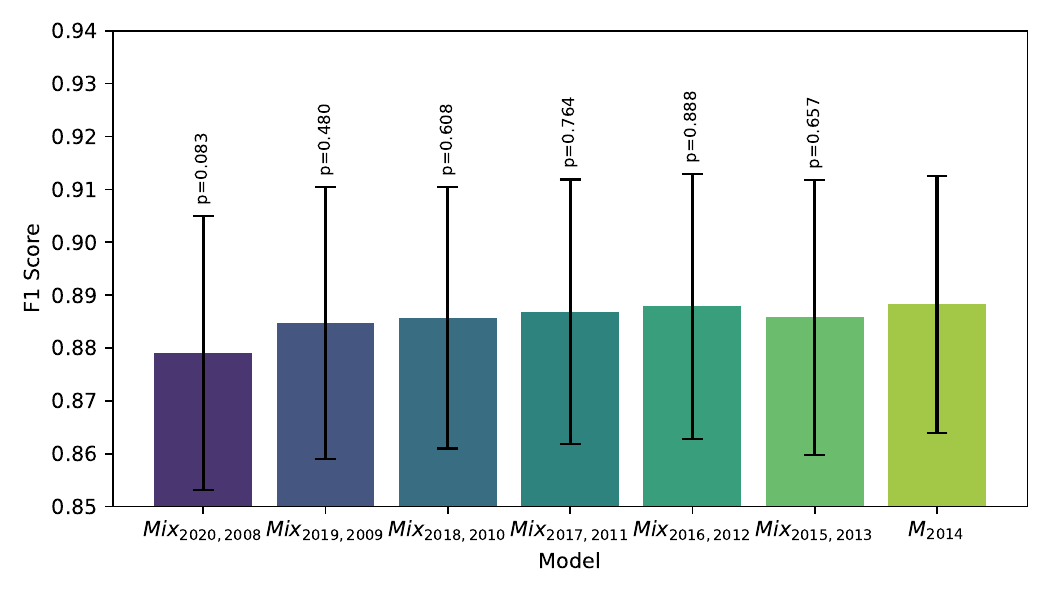}
\caption*{(a)}
\label{fig:interpolate_macro}
\end{minipage}%
\hfill
\begin{minipage}[b]{0.33\textwidth}
\centering
\includegraphics[trim=0 15mm 0 0, width=1\linewidth]{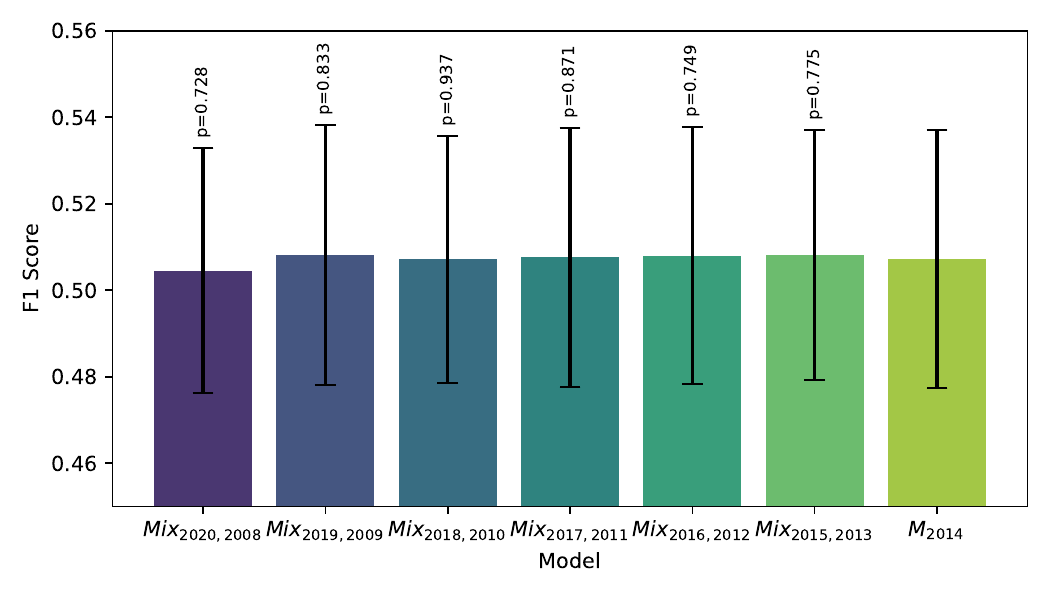}
\caption*{(b)}
\label{fig:interpolate_micro}
\end{minipage}%
\hfill
\begin{minipage}[b]{0.33\textwidth}
\centering
\includegraphics[trim=0 15mm 0 0, width=1\linewidth]{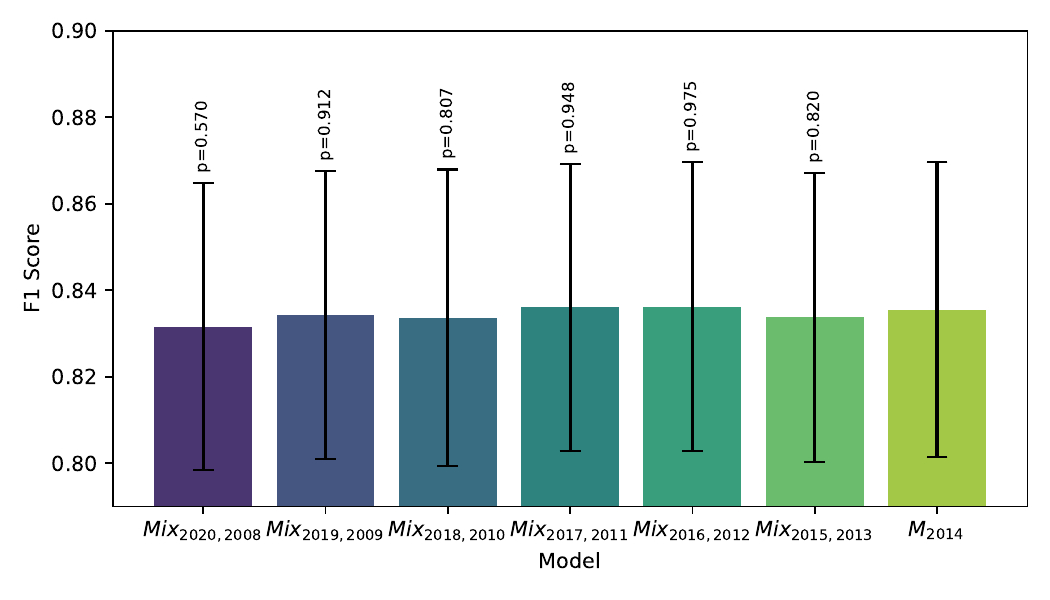}
\caption*{(c)}
\label{fig:interpolate_cross}
\end{minipage}

\caption{Average F1 scores of $\mathcal{M}_{t_0}$ and interpolated models (denoted as $\text{Mix}_{t_1,t_2}$) between $\mathcal{M}_{t_1}$ and $\mathcal{M}_{t_2}$. (a) Major category classification; (b) Subcategory classification; (c) Cross-field identification.}
\label{fig:mix_model_performance}
\end{figure*}

Focusing on $t_0 = 2014$, we have 6 interpolated models by varying $\Delta t$, and Fig.~\ref{fig:mix_model_performance} presents the average F1 scores of these models together with $\mathcal{M}_{2014}$ for the 3 tasks. We also display the $p$-values of the Mann-Whitney U tests comparing F1 scores of each interpolated model with $\mathcal{M}_{2014}$. Firstly, we observe that across the 3 tasks, all the interpolated models perform well and exhibit comparable performances with the real model. To further demonstrate that both $\mathcal{M}_{t_0 - \Delta t}$ and $\mathcal{M}_{t_0 + \Delta t}$ are necessary in creating the interpolated model, we stress that using either of them would not generate similar performance as $\mathcal{M}_{t_0}$ (see Figs.~\ref{fig:heatmaps_classification}a, c, d). Furthermore, we conduct additional experiments where we replace $\mathcal{M}_{2008}$ with a purely random model $\mathcal{M}_{2008}^{\text{random}}$ that has the same mean and variance as $\mathcal{M}_{2008}$, and interpolating between $\mathcal{M}_{2020}$ and $\mathcal{M}_{2008}^{\text{random}}$ yields significantly worse performances than $\mathcal{M}_{2014}$ for the 3 tasks (F1 scores are 0.66, 0.07, and 0.79 respectively, with all $p$-values below 0.05.). 

The second observation from Fig.~\ref{fig:mix_model_performance} is that as $\Delta t$ increases (interpolating between two models farther way temporarily), the performance gap between $\mathcal{M}_{2014}^{\Delta}$ and $\mathcal{M}_{2014}$ tend to increase, although lack of statistical significance. This may imply that the model weights encode temporal information via continual training, and it is feasible to edit the model to fulfill the desired task within a specific time frame. 

\section{Conclusion}

In this work, we presented a new method for pretraining a series of models by organizing the corpora chronologically and introduced AnnualBERT, a series of language models for academic text with different years. This series of language models allows for the tracking of scientific documents across various time periods, facilitating the representation of knowledge over time. We validated our models by fine-tuning them on several standard benchmark datasets, demonstrating that our method achieved a comparable performance to similar models. Moreover, we employed domain-specific link prediction tasks on the arXiv dataset to evaluate the performance of our model, and the results confirmed that our model achieved the state of the art performance in the domain-specific context. To comprehend how models adjust to the evolution of scientific discourse over extended periods, we visualized the model weights and employed a probing task classifying the arXiv corpus to quantify the model's behavior in terms of learning and forgetting as time progresses. Our findings revealed that our model not only exhibits performance on par with the other domain-specific BERT models in standard natural language processing tasks but also captures and reflects the unique linguistic characteristics of the corresponding publication year.

\begin{acks}
This work is supported by the National Natural Science Foundation of China (72204206), City University of Hong Kong (Project No. 9610552, 7005968), and the Hong Kong Institute for Data Science. 
\end{acks}


\providecommand{\noopsort}[1]{}\providecommand{\singleletter}[1]{#1}%


\begin{thebibliography}{58}


\ifx \showCODEN    \undefined \def \showCODEN     #1{\unskip}     \fi
\ifx \showISBNx    \undefined \def \showISBNx     #1{\unskip}     \fi
\ifx \showISBNxiii \undefined \def \showISBNxiii  #1{\unskip}     \fi
\ifx \showISSN     \undefined \def \showISSN      #1{\unskip}     \fi
\ifx \showLCCN     \undefined \def \showLCCN      #1{\unskip}     \fi
\ifx \shownote     \undefined \def \shownote      #1{#1}          \fi
\ifx \showarticletitle \undefined \def \showarticletitle #1{#1}   \fi
\ifx \showURL      \undefined \def \showURL       {\relax}        \fi
\providecommand\bibfield[2]{#2}
\providecommand\bibinfo[2]{#2}
\providecommand\natexlab[1]{#1}
\providecommand\showeprint[2][]{arXiv:#2}

\bibitem[Ahmad et~al\mbox{.}(2022)]%
        {ahmad2022ChemBERTa2}
\bibfield{author}{\bibinfo{person}{Walid Ahmad}, \bibinfo{person}{Elana Simon},
  \bibinfo{person}{Seyone Chithrananda}, \bibinfo{person}{Gabriel Grand}, {and}
  \bibinfo{person}{Bharath Ramsundar}.} \bibinfo{year}{2022}\natexlab{}.
\newblock \showarticletitle{Chemberta-2: Towards chemical foundation models}.
\newblock \bibinfo{journal}{\emph{arXiv preprint arXiv:2209.01712}}
  (\bibinfo{year}{2022}).
\newblock


\bibitem[Anderson et~al\mbox{.}(2012)]%
        {anderson2012towards}
\bibfield{author}{\bibinfo{person}{Ashton Anderson}, \bibinfo{person}{Dan
  Jurafsky}, {and} \bibinfo{person}{Dan McFarland}.}
  \bibinfo{year}{2012}\natexlab{}.
\newblock \showarticletitle{Towards a computational history of the ACL:
  1980-2008}. In \bibinfo{booktitle}{\emph{Proceedings of the ACL-2012 Special
  Workshop on Rediscovering 50 Years of Discoveries}}. \bibinfo{pages}{13--21}.
\newblock


\bibitem[Augenstein et~al\mbox{.}(2017)]%
        {augenstein2017semeval}
\bibfield{author}{\bibinfo{person}{Isabelle Augenstein},
  \bibinfo{person}{Mrinal Das}, \bibinfo{person}{Sebastian Riedel},
  \bibinfo{person}{Lakshmi Vikraman}, {and} \bibinfo{person}{Andrew McCallum}.}
  \bibinfo{year}{2017}\natexlab{}.
\newblock \showarticletitle{Semeval 2017 task 10: Scienceie-extracting
  keyphrases and relations from scientific publications}.
\newblock \bibinfo{journal}{\emph{arXiv preprint arXiv:1704.02853}}
  (\bibinfo{year}{2017}).
\newblock


\bibitem[Beltagy et~al\mbox{.}(2019)]%
        {beltagy2019scibert}
\bibfield{author}{\bibinfo{person}{Iz Beltagy}, \bibinfo{person}{Kyle Lo},
  {and} \bibinfo{person}{Arman Cohan}.} \bibinfo{year}{2019}\natexlab{}.
\newblock \showarticletitle{{S}ci{BERT}: A Pretrained Language Model for
  Scientific Text}. In \bibinfo{booktitle}{\emph{Proceedings of the 2019
  Conference on Empirical Methods in Natural Language Processing and the 9th
  International Joint Conference on Natural Language Processing
  (EMNLP-IJCNLP)}}. \bibinfo{pages}{3615--3620}.
\newblock
\href{https://doi.org/10.18653/v1/D19-1371}{doi:\nolinkurl{10.18653/v1/D19-1371}}


\bibitem[Blei and Lafferty(2006)]%
        {blei2006dynamic}
\bibfield{author}{\bibinfo{person}{David~M Blei} {and} \bibinfo{person}{John~D
  Lafferty}.} \bibinfo{year}{2006}\natexlab{}.
\newblock \showarticletitle{Dynamic topic models}. In
  \bibinfo{booktitle}{\emph{Proceedings of the 23rd International Conference on
  Machine Learning}}. \bibinfo{pages}{113--120}.
\newblock
\href{https://doi.org/10.1145/1143844.1143859}{doi:\nolinkurl{10.1145/1143844.1143859}}


\bibitem[Bolton et~al\mbox{.}(2024)]%
        {bolton2024biomedlm}
\bibfield{author}{\bibinfo{person}{Elliot Bolton}, \bibinfo{person}{Abhinav
  Venigalla}, \bibinfo{person}{Michihiro Yasunaga}, \bibinfo{person}{David
  Hall}, \bibinfo{person}{Betty Xiong}, \bibinfo{person}{Tony Lee},
  \bibinfo{person}{Roxana Daneshjou}, \bibinfo{person}{Jonathan Frankle},
  \bibinfo{person}{Percy Liang}, \bibinfo{person}{Michael Carbin},
  {et~al\mbox{.}}} \bibinfo{year}{2024}\natexlab{}.
\newblock \showarticletitle{Biomedlm: A 2.7 b parameter language model trained
  on biomedical text}.
\newblock \bibinfo{journal}{\emph{arXiv preprint arXiv:2403.18421}}
  (\bibinfo{year}{2024}).
\newblock


\bibitem[Cohan et~al\mbox{.}(2019)]%
        {cohan-etal-2019-structural}
\bibfield{author}{\bibinfo{person}{Arman Cohan}, \bibinfo{person}{Waleed
  Ammar}, \bibinfo{person}{Madeleine van Zuylen}, {and} \bibinfo{person}{Field
  Cady}.} \bibinfo{year}{2019}\natexlab{}.
\newblock \showarticletitle{Structural Scaffolds for Citation Intent
  Classification in Scientific Publications}. In
  \bibinfo{booktitle}{\emph{Proceedings of the 2019 Conference of the North
  {A}merican Chapter of the Association for Computational Linguistics: Human
  Language Technologies, Volume 1 (Long and Short Papers)}}.
  \bibinfo{pages}{3586--3596}.
\newblock
\href{https://doi.org/10.18653/v1/N19-1361}{doi:\nolinkurl{10.18653/v1/N19-1361}}


\bibitem[Devlin et~al\mbox{.}(2018)]%
        {devlin2018bert}
\bibfield{author}{\bibinfo{person}{Jacob Devlin}, \bibinfo{person}{Ming-Wei
  Chang}, \bibinfo{person}{Kenton Lee}, {and} \bibinfo{person}{Kristina
  Toutanova}.} \bibinfo{year}{2018}\natexlab{}.
\newblock \showarticletitle{Bert: Pre-training of deep bidirectional
  transformers for language understanding}.
\newblock \bibinfo{journal}{\emph{arXiv preprint arXiv:1810.04805}}
  (\bibinfo{year}{2018}).
\newblock


\bibitem[Dhingra et~al\mbox{.}(2022)]%
        {dhingra2022time}
\bibfield{author}{\bibinfo{person}{Bhuwan Dhingra}, \bibinfo{person}{Jeremy~R.
  Cole}, \bibinfo{person}{Julian~Martin Eisenschlos}, \bibinfo{person}{Daniel
  Gillick}, \bibinfo{person}{Jacob Eisenstein}, {and}
  \bibinfo{person}{William~W. Cohen}.} \bibinfo{year}{2022}\natexlab{}.
\newblock \showarticletitle{Time-Aware Language Models as Temporal Knowledge
  Bases}.
\newblock \bibinfo{journal}{\emph{Transactions of the Association for
  Computational Linguistics}}  \bibinfo{volume}{10} (\bibinfo{year}{2022}),
  \bibinfo{pages}{257--273}.
\newblock
\href{https://doi.org/10.1162/tacl_a_00459}{doi:\nolinkurl{10.1162/tacl_a_00459}}


\bibitem[Elnaggar et~al\mbox{.}(2021)]%
        {elnaggar2021prottrans}
\bibfield{author}{\bibinfo{person}{Ahmed Elnaggar}, \bibinfo{person}{Michael
  Heinzinger}, \bibinfo{person}{Christian Dallago}, \bibinfo{person}{Ghalia
  Rehawi}, \bibinfo{person}{Yu Wang}, \bibinfo{person}{Llion Jones},
  \bibinfo{person}{Tom Gibbs}, \bibinfo{person}{Tamas Feher},
  \bibinfo{person}{Christoph Angerer}, \bibinfo{person}{Martin Steinegger},
  {et~al\mbox{.}}} \bibinfo{year}{2021}\natexlab{}.
\newblock \showarticletitle{Prottrans: Toward understanding the language of
  life through self-supervised learning}.
\newblock \bibinfo{journal}{\emph{IEEE Transactions on Pattern Analysis and
  Machine Intelligence}} \bibinfo{volume}{44}, \bibinfo{number}{10}
  (\bibinfo{year}{2021}), \bibinfo{pages}{7112--7127}.
\newblock
\href{https://doi.org/10.1109/TPAMI.2021.3095381}{doi:\nolinkurl{10.1109/TPAMI.2021.3095381}}


\bibitem[Fan et~al\mbox{.}(2024)]%
        {fan2024generalizing}
\bibfield{author}{\bibinfo{person}{Shaohua Fan}, \bibinfo{person}{Xiao Wang},
  \bibinfo{person}{Chuan Shi}, \bibinfo{person}{Peng Cui}, {and}
  \bibinfo{person}{Bai Wang}.} \bibinfo{year}{2024}\natexlab{}.
\newblock \showarticletitle{Generalizing graph neural networks on
  out-of-distribution graphs}.
\newblock \bibinfo{journal}{\emph{IEEE Transactions on Pattern Analysis and
  Machine Intelligence}} \bibinfo{volume}{46}, \bibinfo{number}{1}
  (\bibinfo{year}{2024}), \bibinfo{pages}{322--337}.
\newblock
\href{https://doi.org/10.1109/TPAMI.2023.3321097}{doi:\nolinkurl{10.1109/TPAMI.2023.3321097}}


\bibitem[Feng et~al\mbox{.}(2024)]%
        {feng2024pretraining}
\bibfield{author}{\bibinfo{person}{Zhangyin Feng}, \bibinfo{person}{Duyu Tang},
  \bibinfo{person}{Xiaocheng Feng}, \bibinfo{person}{Cong Zhou},
  \bibinfo{person}{Junwei Liao}, \bibinfo{person}{Shuangzhi Wu},
  \bibinfo{person}{Bing Qin}, \bibinfo{person}{Yunbo Cao}, {and}
  \bibinfo{person}{Shuming Shi}.} \bibinfo{year}{2024}\natexlab{}.
\newblock \showarticletitle{Pretraining without wordpieces: learning over a
  vocabulary of millions of words}.
\newblock \bibinfo{journal}{\emph{International Journal of Machine Learning and
  Cybernetics}} (\bibinfo{year}{2024}), \bibinfo{pages}{1--10}.
\newblock
\href{https://doi.org/10.1007/s13042-024-02132-4}{doi:\nolinkurl{10.1007/s13042-024-02132-4}}


\bibitem[Gates et~al\mbox{.}(2019)]%
        {gates2019nature}
\bibfield{author}{\bibinfo{person}{Alexander~J. Gates}, \bibinfo{person}{Qing
  Ke}, \bibinfo{person}{Onur Varol}, {and} \bibinfo{person}{Albert-László
  Barabási}.} \bibinfo{year}{2019}\natexlab{}.
\newblock \showarticletitle{Nature's reach: narrow work has broad impact}.
\newblock \bibinfo{journal}{\emph{Nature}}  \bibinfo{volume}{575}
  (\bibinfo{year}{2019}), \bibinfo{pages}{32--34}.
\newblock
\href{https://doi.org/10.1038/d41586-019-03308-7}{doi:\nolinkurl{10.1038/d41586-019-03308-7}}


\bibitem[Ghasemian et~al\mbox{.}(2020)]%
        {ghasemian2020stacking}
\bibfield{author}{\bibinfo{person}{Amir Ghasemian}, \bibinfo{person}{Homa
  Hosseinmardi}, \bibinfo{person}{Aram Galstyan}, \bibinfo{person}{Edoardo~M
  Airoldi}, {and} \bibinfo{person}{Aaron Clauset}.}
  \bibinfo{year}{2020}\natexlab{}.
\newblock \showarticletitle{Stacking models for nearly optimal link prediction
  in complex networks}.
\newblock \bibinfo{journal}{\emph{Proceedings of the National Academy of
  Sciences}} \bibinfo{volume}{117}, \bibinfo{number}{38}
  (\bibinfo{year}{2020}), \bibinfo{pages}{23393--23400}.
\newblock
\href{https://doi.org/10.1073/pnas.1914950117}{doi:\nolinkurl{10.1073/pnas.1914950117}}


\bibitem[Gu et~al\mbox{.}(2021)]%
        {gu2021domain}
\bibfield{author}{\bibinfo{person}{Yu Gu}, \bibinfo{person}{Robert Tinn},
  \bibinfo{person}{Hao Cheng}, \bibinfo{person}{Michael Lucas},
  \bibinfo{person}{Naoto Usuyama}, \bibinfo{person}{Xiaodong Liu},
  \bibinfo{person}{Tristan Naumann}, \bibinfo{person}{Jianfeng Gao}, {and}
  \bibinfo{person}{Hoifung Poon}.} \bibinfo{year}{2021}\natexlab{}.
\newblock \showarticletitle{Domain-specific language model pretraining for
  biomedical natural language processing}.
\newblock \bibinfo{journal}{\emph{ACM Transactions on Computing for Healthcare
  (HEALTH)}} \bibinfo{volume}{3}, \bibinfo{number}{1} (\bibinfo{year}{2021}),
  \bibinfo{pages}{1--23}.
\newblock
\href{https://doi.org/10.1145/3458754}{doi:\nolinkurl{10.1145/3458754}}


\bibitem[Guo et~al\mbox{.}(2021)]%
        {guo2021automated}
\bibfield{author}{\bibinfo{person}{Jiang Guo}, \bibinfo{person}{A~Santiago
  Ibanez-Lopez}, \bibinfo{person}{Hanyu Gao}, \bibinfo{person}{Victor Quach},
  \bibinfo{person}{Connor~W Coley}, \bibinfo{person}{Klavs~F Jensen}, {and}
  \bibinfo{person}{Regina Barzilay}.} \bibinfo{year}{2021}\natexlab{}.
\newblock \showarticletitle{Automated chemical reaction extraction from
  scientific literature}.
\newblock \bibinfo{journal}{\emph{Journal of Chemical Information and
  Modeling}} \bibinfo{volume}{62}, \bibinfo{number}{9} (\bibinfo{year}{2021}),
  \bibinfo{pages}{2035--2045}.
\newblock
\href{https://doi.org/10.1021/acs.jcim.1c00284}{doi:\nolinkurl{10.1021/acs.jcim.1c00284}}


\bibitem[Gurnee and Tegmark(2023)]%
        {gurnee2023language}
\bibfield{author}{\bibinfo{person}{Wes Gurnee} {and} \bibinfo{person}{Max
  Tegmark}.} \bibinfo{year}{2023}\natexlab{}.
\newblock \showarticletitle{Language models represent space and time}.
\newblock \bibinfo{journal}{\emph{arXiv preprint arXiv:2310.02207}}
  (\bibinfo{year}{2023}).
\newblock


\bibitem[Gururangan et~al\mbox{.}(2020)]%
        {gururangan2020dont}
\bibfield{author}{\bibinfo{person}{Suchin Gururangan}, \bibinfo{person}{Ana
  Marasovi{\'c}}, \bibinfo{person}{Swabha Swayamdipta}, \bibinfo{person}{Kyle
  Lo}, \bibinfo{person}{Iz Beltagy}, \bibinfo{person}{Doug Downey}, {and}
  \bibinfo{person}{Noah~A. Smith}.} \bibinfo{year}{2020}\natexlab{}.
\newblock \showarticletitle{Don{'}t Stop Pretraining: Adapt Language Models to
  Domains and Tasks}. In \bibinfo{booktitle}{\emph{Proceedings of the 58th
  Annual Meeting of the Association for Computational Linguistics}}.
  \bibinfo{pages}{8342--8360}.
\newblock
\href{https://doi.org/10.18653/v1/2020.acl-main.740}{doi:\nolinkurl{10.18653/v1/2020.acl-main.740}}


\bibitem[Hamilton et~al\mbox{.}(2017)]%
        {Hamilton2017InductiveRL}
\bibfield{author}{\bibinfo{person}{Will Hamilton}, \bibinfo{person}{Zhitao
  Ying}, {and} \bibinfo{person}{Jure Leskovec}.}
  \bibinfo{year}{2017}\natexlab{}.
\newblock \showarticletitle{Inductive Representation Learning on Large Graphs}.
  In \bibinfo{booktitle}{\emph{Advances in Neural Information Processing
  Systems}}, Vol.~\bibinfo{volume}{30}.
\newblock


\bibitem[Ho et~al\mbox{.}(2024)]%
        {ho2024survey}
\bibfield{author}{\bibinfo{person}{Xanh Ho}, \bibinfo{person}{Anh Khoa~Duong
  Nguyen}, \bibinfo{person}{An~Tuan Dao}, \bibinfo{person}{Junfeng Jiang},
  \bibinfo{person}{Yuki Chida}, \bibinfo{person}{Kaito Sugimoto},
  \bibinfo{person}{Huy~Quoc To}, \bibinfo{person}{Florian Boudin}, {and}
  \bibinfo{person}{Akiko Aizawa}.} \bibinfo{year}{2024}\natexlab{}.
\newblock \showarticletitle{A Survey of Pre-trained Language Models for
  Processing Scientific Text}.
\newblock \bibinfo{journal}{\emph{arXiv preprint arXiv:2401.17824}}
  (\bibinfo{year}{2024}).
\newblock


\bibitem[Hong et~al\mbox{.}(2023)]%
        {hong2023diminishing}
\bibfield{author}{\bibinfo{person}{Zhi Hong}, \bibinfo{person}{Aswathy Ajith},
  \bibinfo{person}{James Pauloski}, \bibinfo{person}{Eamon Duede},
  \bibinfo{person}{Kyle Chard}, {and} \bibinfo{person}{Ian Foster}.}
  \bibinfo{year}{2023}\natexlab{}.
\newblock \showarticletitle{The Diminishing Returns of Masked Language Models
  to Science}. In \bibinfo{booktitle}{\emph{Findings of the Association for
  Computational Linguistics: ACL 2023}}. \bibinfo{pages}{1270--1283}.
\newblock
\href{https://doi.org/10.18653/v1/2023.findings-acl.82}{doi:\nolinkurl{10.18653/v1/2023.findings-acl.82}}


\bibitem[Huang et~al\mbox{.}(2024)]%
        {huang2024self}
\bibfield{author}{\bibinfo{person}{Lang Huang}, \bibinfo{person}{Chao Zhang},
  {and} \bibinfo{person}{Hongyang Zhang}.} \bibinfo{year}{2024}\natexlab{}.
\newblock \showarticletitle{Self-adaptive training: Bridging supervised and
  self-supervised learning}.
\newblock \bibinfo{journal}{\emph{IEEE Transactions on Pattern Analysis and
  Machine Intelligence}} \bibinfo{volume}{46}, \bibinfo{number}{3}
  (\bibinfo{year}{2024}), \bibinfo{pages}{1362--1377}.
\newblock
\href{https://doi.org/10.1109/TPAMI.2022.3217792}{doi:\nolinkurl{10.1109/TPAMI.2022.3217792}}


\bibitem[Ilharco et~al\mbox{.}(2023)]%
        {ilharco2022editing}
\bibfield{author}{\bibinfo{person}{Gabriel Ilharco},
  \bibinfo{person}{Marco~Tulio Ribeiro}, \bibinfo{person}{Mitchell Wortsman},
  \bibinfo{person}{Suchin Gururangan}, \bibinfo{person}{Ludwig Schmidt},
  \bibinfo{person}{Hannaneh Hajishirzi}, {and} \bibinfo{person}{Ali Farhadi}.}
  \bibinfo{year}{2023}\natexlab{}.
\newblock \showarticletitle{Editing models with task arithmetic}. In
  \bibinfo{booktitle}{\emph{Proceedings of the 11th International Conference on
  Learning Representations}}.
\newblock


\bibitem[Jurgens et~al\mbox{.}(2018)]%
        {jurgens2018measuring}
\bibfield{author}{\bibinfo{person}{David Jurgens}, \bibinfo{person}{Srijan
  Kumar}, \bibinfo{person}{Raine Hoover}, \bibinfo{person}{Dan McFarland},
  {and} \bibinfo{person}{Dan Jurafsky}.} \bibinfo{year}{2018}\natexlab{}.
\newblock \showarticletitle{Measuring the evolution of a scientific field
  through citation frames}.
\newblock \bibinfo{journal}{\emph{Transactions of the Association for
  Computational Linguistics}}  \bibinfo{volume}{6} (\bibinfo{year}{2018}),
  \bibinfo{pages}{391--406}.
\newblock
\href{https://doi.org/10.1162/tacl_a_00028}{doi:\nolinkurl{10.1162/tacl_a_00028}}


\bibitem[Kanakarajan et~al\mbox{.}(2021)]%
        {kanakarajan2021BioELECTRA}
\bibfield{author}{\bibinfo{person}{Kamal~Raj Kanakarajan},
  \bibinfo{person}{Bhuvana Kundumani}, {and} \bibinfo{person}{Malaikannan
  Sankarasubbu}.} \bibinfo{year}{2021}\natexlab{}.
\newblock \showarticletitle{BioELECTRA: pretrained biomedical text encoder
  using discriminators}. In \bibinfo{booktitle}{\emph{Proceedings of the 20th
  Workshop on Biomedical Language Processing}}. \bibinfo{pages}{143--154}.
\newblock
\href{https://doi.org/10.18653/v1/2021.bionlp-1.16}{doi:\nolinkurl{10.18653/v1/2021.bionlp-1.16}}


\bibitem[Ke(2020)]%
        {ke2020technological}
\bibfield{author}{\bibinfo{person}{Qing Ke}.} \bibinfo{year}{2020}\natexlab{}.
\newblock \showarticletitle{Technological impact of biomedical research: The
  role of basicness and novelty}.
\newblock \bibinfo{journal}{\emph{Research Policy}} \bibinfo{volume}{49},
  \bibinfo{number}{7} (\bibinfo{year}{2020}), \bibinfo{pages}{104071}.
\newblock
\href{https://doi.org/10.1016/j.respol.2020.104071}{doi:\nolinkurl{10.1016/j.respol.2020.104071}}


\bibitem[Lazaridou et~al\mbox{.}(2021)]%
        {lazaridou2021mind}
\bibfield{author}{\bibinfo{person}{Angeliki Lazaridou}, \bibinfo{person}{Adhi
  Kuncoro}, \bibinfo{person}{Elena Gribovskaya}, \bibinfo{person}{Devang
  Agrawal}, \bibinfo{person}{Adam Liska}, \bibinfo{person}{Tayfun Terzi},
  \bibinfo{person}{Mai Gimenez}, \bibinfo{person}{Cyprien de Masson~d'Autume},
  \bibinfo{person}{Tomas Kocisky}, \bibinfo{person}{Sebastian Ruder},
  {et~al\mbox{.}}} \bibinfo{year}{2021}\natexlab{}.
\newblock \showarticletitle{Mind the gap: Assessing temporal generalization in
  neural language models}.
\newblock \bibinfo{journal}{\emph{Advances in Neural Information Processing
  Systems}}  \bibinfo{volume}{34} (\bibinfo{year}{2021}),
  \bibinfo{pages}{29348--29363}.
\newblock


\bibitem[Ledford(2015)]%
        {ledford2015how}
\bibfield{author}{\bibinfo{person}{Heidi Ledford}.}
  \bibinfo{year}{2015}\natexlab{}.
\newblock \showarticletitle{How to solve the world's biggest problems}.
\newblock \bibinfo{journal}{\emph{Nature}} \bibinfo{volume}{525},
  \bibinfo{number}{7569} (\bibinfo{year}{2015}), \bibinfo{pages}{308--311}.
\newblock
\href{https://doi.org/10.1038/525308a}{doi:\nolinkurl{10.1038/525308a}}


\bibitem[Lee et~al\mbox{.}(2020)]%
        {lee2020biobert}
\bibfield{author}{\bibinfo{person}{Jinhyuk Lee}, \bibinfo{person}{Wonjin Yoon},
  \bibinfo{person}{Sungdong Kim}, \bibinfo{person}{Donghyeon Kim},
  \bibinfo{person}{Sunkyu Kim}, \bibinfo{person}{Chan~Ho So}, {and}
  \bibinfo{person}{Jaewoo Kang}.} \bibinfo{year}{2020}\natexlab{}.
\newblock \showarticletitle{BioBERT: a pre-trained biomedical language
  representation model for biomedical text mining}.
\newblock \bibinfo{journal}{\emph{Bioinformatics}} \bibinfo{volume}{36},
  \bibinfo{number}{4} (\bibinfo{year}{2020}), \bibinfo{pages}{1234--1240}.
\newblock
\href{https://doi.org/10.1093/bioinformatics/btz682}{doi:\nolinkurl{10.1093/bioinformatics/btz682}}


\bibitem[Liben-Nowell and Kleinberg(2007)]%
        {liben2007link}
\bibfield{author}{\bibinfo{person}{David Liben-Nowell} {and}
  \bibinfo{person}{Jon Kleinberg}.} \bibinfo{year}{2007}\natexlab{}.
\newblock \showarticletitle{The link prediction problem for social networks}.
\newblock \bibinfo{journal}{\emph{Journal of the American Society for
  Information Science and Technology}}  \bibinfo{volume}{58}
  (\bibinfo{year}{2007}), \bibinfo{pages}{1019--1031}.
\newblock
Issue 7.
\href{https://doi.org/10.1002/asi.20591}{doi:\nolinkurl{10.1002/asi.20591}}


\bibitem[Liu et~al\mbox{.}(2022b)]%
        {liu2022oag}
\bibfield{author}{\bibinfo{person}{Xiao Liu}, \bibinfo{person}{Da Yin},
  \bibinfo{person}{Jingnan Zheng}, \bibinfo{person}{Xingjian Zhang},
  \bibinfo{person}{Peng Zhang}, \bibinfo{person}{Hongxia Yang},
  \bibinfo{person}{Yuxiao Dong}, {and} \bibinfo{person}{Jie Tang}.}
  \bibinfo{year}{2022}\natexlab{b}.
\newblock \showarticletitle{Oag-bert: Towards a unified backbone language model
  for academic knowledge services}. In \bibinfo{booktitle}{\emph{Proceedings of
  the 28th ACM SIGKDD Conference on Knowledge Discovery and Data Mining}}.
  \bibinfo{pages}{3418--3428}.
\newblock
\href{https://doi.org/10.1145/3534678.3539210}{doi:\nolinkurl{10.1145/3534678.3539210}}


\bibitem[Liu et~al\mbox{.}(2022a)]%
        {liu2022learning}
\bibfield{author}{\bibinfo{person}{Yufan Liu}, \bibinfo{person}{Jiajiong Cao},
  \bibinfo{person}{Bing Li}, \bibinfo{person}{Weiming Hu}, {and}
  \bibinfo{person}{Stephen Maybank}.} \bibinfo{year}{2022}\natexlab{a}.
\newblock \showarticletitle{Learning to explore distillability and
  sparsability: a joint framework for model compression}.
\newblock \bibinfo{journal}{\emph{IEEE Transactions on Pattern Analysis and
  Machine Intelligence}} \bibinfo{volume}{45}, \bibinfo{number}{3}
  (\bibinfo{year}{2022}), \bibinfo{pages}{3378--3395}.
\newblock


\bibitem[Liu et~al\mbox{.}(2019)]%
        {liu2019roberta}
\bibfield{author}{\bibinfo{person}{Yinhan Liu}, \bibinfo{person}{Myle Ott},
  \bibinfo{person}{Naman Goyal}, \bibinfo{person}{Jingfei Du},
  \bibinfo{person}{Mandar Joshi}, \bibinfo{person}{Danqi Chen},
  \bibinfo{person}{Omer Levy}, \bibinfo{person}{Mike Lewis},
  \bibinfo{person}{Luke Zettlemoyer}, {and} \bibinfo{person}{Veselin
  Stoyanov}.} \bibinfo{year}{2019}\natexlab{}.
\newblock \showarticletitle{Roberta: A robustly optimized bert pretraining
  approach}.
\newblock \bibinfo{journal}{\emph{arXiv preprint arXiv:1907.11692}}
  (\bibinfo{year}{2019}).
\newblock


\bibitem[Lo et~al\mbox{.}(2020)]%
        {s2orc}
\bibfield{author}{\bibinfo{person}{Kyle Lo}, \bibinfo{person}{Lucy~Lu Wang},
  \bibinfo{person}{Mark Neumann}, \bibinfo{person}{Rodney Kinney}, {and}
  \bibinfo{person}{Daniel~S Weld}.} \bibinfo{year}{2020}\natexlab{}.
\newblock \showarticletitle{S2ORC: The Semantic Scholar Open Research Corpus}.
  In \bibinfo{booktitle}{\emph{Proceedings of the 58th Annual Meeting of the
  Association for Computational Linguistics}}. \bibinfo{pages}{4969--4983}.
\newblock


\bibitem[Loper and Bird(2002)]%
        {loper2002nltk}
\bibfield{author}{\bibinfo{person}{Edward Loper} {and} \bibinfo{person}{Steven
  Bird}.} \bibinfo{year}{2002}\natexlab{}.
\newblock \showarticletitle{Nltk: The natural language toolkit}.
\newblock \bibinfo{journal}{\emph{arXiv preprint cs/0205028}}
  (\bibinfo{year}{2002}).
\newblock


\bibitem[Luan et~al\mbox{.}(2018)]%
        {luan-etal-2018-multi}
\bibfield{author}{\bibinfo{person}{Yi Luan}, \bibinfo{person}{Luheng He},
  \bibinfo{person}{Mari Ostendorf}, {and} \bibinfo{person}{Hannaneh
  Hajishirzi}.} \bibinfo{year}{2018}\natexlab{}.
\newblock \showarticletitle{Multi-Task Identification of Entities, Relations,
  and Coreference for Scientific Knowledge Graph Construction}. In
  \bibinfo{booktitle}{\emph{Proceedings of the 2018 Conference on Empirical
  Methods in Natural Language Processing}}. \bibinfo{pages}{3219--3232}.
\newblock
\href{https://doi.org/10.18653/v1/D18-1360}{doi:\nolinkurl{10.18653/v1/D18-1360}}


\bibitem[Luo et~al\mbox{.}(2022)]%
        {luo2022BioGPT}
\bibfield{author}{\bibinfo{person}{Renqian Luo}, \bibinfo{person}{Liai Sun},
  \bibinfo{person}{Yingce Xia}, \bibinfo{person}{Tao Qin},
  \bibinfo{person}{Sheng Zhang}, \bibinfo{person}{Hoifung Poon}, {and}
  \bibinfo{person}{Tie-Yan Liu}.} \bibinfo{year}{2022}\natexlab{}.
\newblock \showarticletitle{BioGPT: generative pre-trained transformer for
  biomedical text generation and mining}.
\newblock \bibinfo{journal}{\emph{Briefings in Bioinformatics}}
  \bibinfo{volume}{23}, \bibinfo{number}{6} (\bibinfo{year}{2022}),
  \bibinfo{pages}{bbac409}.
\newblock
\href{https://doi.org/10.1093/bib/bbac409}{doi:\nolinkurl{10.1093/bib/bbac409}}


\bibitem[Luu et~al\mbox{.}(2022)]%
        {luu2022time}
\bibfield{author}{\bibinfo{person}{Kelvin Luu}, \bibinfo{person}{Daniel
  Khashabi}, \bibinfo{person}{Suchin Gururangan}, \bibinfo{person}{Karishma
  Mandyam}, {and} \bibinfo{person}{Noah~A Smith}.}
  \bibinfo{year}{2022}\natexlab{}.
\newblock \showarticletitle{Time Waits for No One! Analysis and Challenges of
  Temporal Misalignment}. In \bibinfo{booktitle}{\emph{Proceedings of the 2022
  Conference of the North American Chapter of the Association for Computational
  Linguistics: Human Language Technologies}}. \bibinfo{pages}{5944--5958}.
\newblock
\href{https://doi.org/10.18653/v1/2022.naacl-main.435}{doi:\nolinkurl{10.18653/v1/2022.naacl-main.435}}


\bibitem[Matena and Raffel(2022)]%
        {matena2022merging}
\bibfield{author}{\bibinfo{person}{Michael~S Matena} {and}
  \bibinfo{person}{Colin~A Raffel}.} \bibinfo{year}{2022}\natexlab{}.
\newblock \showarticletitle{Merging models with fisher-weighted averaging}.
\newblock \bibinfo{journal}{\emph{Advances in Neural Information Processing
  Systems}}  \bibinfo{volume}{35} (\bibinfo{year}{2022}),
  \bibinfo{pages}{17703--17716}.
\newblock


\bibitem[Miller(2024)]%
        {LaTeXML}
\bibfield{author}{\bibinfo{person}{Bruce Miller}.}
  \bibinfo{year}{2024}\natexlab{}.
\newblock \bibinfo{title}{{LaTeXML}: A {LaTeX} to {XML}/{HTML}/{MathML}
  Converter}.
\newblock
  \bibinfo{howpublished}{\url{https://math.nist.gov/~BMiller/LaTeXML/}}.
\newblock


\bibitem[Naseem et~al\mbox{.}(2021)]%
        {naseem2021bioalbert}
\bibfield{author}{\bibinfo{person}{Usman Naseem}, \bibinfo{person}{Matloob
  Khushi}, \bibinfo{person}{Vinay Reddy}, \bibinfo{person}{Sakthivel
  Rajendran}, \bibinfo{person}{Imran Razzak}, {and} \bibinfo{person}{Jinman
  Kim}.} \bibinfo{year}{2021}\natexlab{}.
\newblock \showarticletitle{Bioalbert: A simple and effective pre-trained
  language model for biomedical named entity recognition}. In
  \bibinfo{booktitle}{\emph{2021 International Joint Conference on Neural
  Networks (IJCNN)}}. \bibinfo{pages}{1--7}.
\newblock
\href{https://doi.org/10.1109/IJCNN52387.2021.9533884}{doi:\nolinkurl{10.1109/IJCNN52387.2021.9533884}}


\bibitem[Nylund et~al\mbox{.}(2023)]%
        {nylund2023time}
\bibfield{author}{\bibinfo{person}{Kai Nylund}, \bibinfo{person}{Suchin
  Gururangan}, {and} \bibinfo{person}{Noah~A Smith}.}
  \bibinfo{year}{2023}\natexlab{}.
\newblock \showarticletitle{Time is Encoded in the Weights of Finetuned
  Language Models}.
\newblock \bibinfo{journal}{\emph{arXiv preprint arXiv:2312.13401}}
  (\bibinfo{year}{2023}).
\newblock


\bibitem[Peng et~al\mbox{.}(2021a)]%
        {peng2021neural}
\bibfield{author}{\bibinfo{person}{Hao Peng}, \bibinfo{person}{Qing Ke},
  \bibinfo{person}{Ceren Budak}, \bibinfo{person}{Daniel~M. Romero}, {and}
  \bibinfo{person}{Yong-Yeol Ahn}.} \bibinfo{year}{2021}\natexlab{a}.
\newblock \showarticletitle{Neural embeddings of scholarly periodicals reveal
  complex disciplinary organizations}.
\newblock \bibinfo{journal}{\emph{Science Advances}} \bibinfo{volume}{7},
  \bibinfo{number}{17} (\bibinfo{year}{2021}), \bibinfo{pages}{eabb9004}.
\newblock
\href{https://doi.org/10.1126/sciadv.abb9004}{doi:\nolinkurl{10.1126/sciadv.abb9004}}


\bibitem[Peng et~al\mbox{.}(2021b)]%
        {peng2021mathbert}
\bibfield{author}{\bibinfo{person}{Shuai Peng}, \bibinfo{person}{Ke Yuan},
  \bibinfo{person}{Liangcai Gao}, {and} \bibinfo{person}{Zhi Tang}.}
  \bibinfo{year}{2021}\natexlab{b}.
\newblock \showarticletitle{Mathbert: A pre-trained model for mathematical
  formula understanding}.
\newblock \bibinfo{journal}{\emph{arXiv preprint arXiv:2105.00377}}
  (\bibinfo{year}{2021}).
\newblock


\bibitem[Prabhakaran et~al\mbox{.}(2016)]%
        {prabhakaran2016predicting}
\bibfield{author}{\bibinfo{person}{Vinodkumar Prabhakaran},
  \bibinfo{person}{William~L Hamilton}, \bibinfo{person}{Dan McFarland}, {and}
  \bibinfo{person}{Dan Jurafsky}.} \bibinfo{year}{2016}\natexlab{}.
\newblock \showarticletitle{Predicting the rise and fall of scientific topics
  from trends in their rhetorical framing}. In
  \bibinfo{booktitle}{\emph{Proceedings of the 54th Annual Meeting of the
  Association for Computational Linguistics (Volume 1: Long Papers)}}.
  \bibinfo{pages}{1170--1180}.
\newblock
\href{https://doi.org/10.18653/v1/P16-1111}{doi:\nolinkurl{10.18653/v1/P16-1111}}


\bibitem[Reimers and Gurevych(2019)]%
        {reimers2019sentence}
\bibfield{author}{\bibinfo{person}{Nils Reimers} {and} \bibinfo{person}{Iryna
  Gurevych}.} \bibinfo{year}{2019}\natexlab{}.
\newblock \showarticletitle{Sentence-{BERT}: Sentence Embeddings using
  {S}iamese {BERT}-Networks}. In \bibinfo{booktitle}{\emph{Proceedings of the
  2019 Conference on Empirical Methods in Natural Language Processing and the
  9th International Joint Conference on Natural Language Processing
  (EMNLP-IJCNLP)}}. \bibinfo{pages}{3982--3992}.
\newblock
\href{https://doi.org/10.18653/v1/D19-1410}{doi:\nolinkurl{10.18653/v1/D19-1410}}


\bibitem[Rijhwani and Preo{\c{t}}iuc-Pietro(2020)]%
        {rijhwani2020temporally}
\bibfield{author}{\bibinfo{person}{Shruti Rijhwani} {and}
  \bibinfo{person}{Daniel Preo{\c{t}}iuc-Pietro}.}
  \bibinfo{year}{2020}\natexlab{}.
\newblock \showarticletitle{Temporally-informed analysis of named entity
  recognition}. In \bibinfo{booktitle}{\emph{Proceedings of the 58th Annual
  Meeting of the Association for Computational Linguistics}}.
  \bibinfo{pages}{7605--7617}.
\newblock
\href{https://doi.org/10.18653/v1/2020.acl-main.680}{doi:\nolinkurl{10.18653/v1/2020.acl-main.680}}


\bibitem[Ross et~al\mbox{.}(2022)]%
        {ross2022large}
\bibfield{author}{\bibinfo{person}{Jerret Ross}, \bibinfo{person}{Brian
  Belgodere}, \bibinfo{person}{Vijil Chenthamarakshan}, \bibinfo{person}{Inkit
  Padhi}, \bibinfo{person}{Youssef Mroueh}, {and} \bibinfo{person}{Payel Das}.}
  \bibinfo{year}{2022}\natexlab{}.
\newblock \showarticletitle{Large-scale chemical language representations
  capture molecular structure and properties}.
\newblock \bibinfo{journal}{\emph{Nature Machine Intelligence}}
  \bibinfo{volume}{4}, \bibinfo{number}{12} (\bibinfo{year}{2022}),
  \bibinfo{pages}{1256--1264}.
\newblock
\href{https://doi.org/10.1038/s42256-022-00580-7}{doi:\nolinkurl{10.1038/s42256-022-00580-7}}


\bibitem[Shen et~al\mbox{.}(2023)]%
        {shen2023sscibert}
\bibfield{author}{\bibinfo{person}{Si Shen}, \bibinfo{person}{Jiangfeng Liu},
  \bibinfo{person}{Litao Lin}, \bibinfo{person}{Ying Huang},
  \bibinfo{person}{Lin Zhang}, \bibinfo{person}{Chang Liu},
  \bibinfo{person}{Yutong Feng}, {and} \bibinfo{person}{Dongbo Wang}.}
  \bibinfo{year}{2023}\natexlab{}.
\newblock \showarticletitle{SsciBERT: A pre-trained language model for social
  science texts}.
\newblock \bibinfo{journal}{\emph{Scientometrics}} \bibinfo{volume}{128},
  \bibinfo{number}{2} (\bibinfo{year}{2023}), \bibinfo{pages}{1241--1263}.
\newblock
\href{https://doi.org/10.1007/s11192-022-04602-4}{doi:\nolinkurl{10.1007/s11192-022-04602-4}}


\bibitem[Sinha et~al\mbox{.}(2015)]%
        {sinha2015overview}
\bibfield{author}{\bibinfo{person}{Arnab Sinha}, \bibinfo{person}{Zhihong
  Shen}, \bibinfo{person}{Yang Song}, \bibinfo{person}{Hao Ma},
  \bibinfo{person}{Darrin Eide}, \bibinfo{person}{Bo-June Hsu}, {and}
  \bibinfo{person}{Kuansan Wang}.} \bibinfo{year}{2015}\natexlab{}.
\newblock \showarticletitle{An overview of microsoft academic service (mas) and
  applications}. In \bibinfo{booktitle}{\emph{Proceedings of the 24th
  International Conference on World Wide Web}}. \bibinfo{pages}{243--246}.
\newblock


\bibitem[Smith et~al\mbox{.}(2020)]%
        {smith-etal-2020-scienceexamcer}
\bibfield{author}{\bibinfo{person}{Hannah Smith}, \bibinfo{person}{Zeyu Zhang},
  \bibinfo{person}{John Culnan}, {and} \bibinfo{person}{Peter Jansen}.}
  \bibinfo{year}{2020}\natexlab{}.
\newblock \showarticletitle{ScienceExamCER: A High-Density Fine-Grained
  Science-Domain Corpus for Common Entity Recognition}. In
  \bibinfo{booktitle}{\emph{Proceedings of the Twelfth Language Resources and
  Evaluation Conference}}. \bibinfo{pages}{4529--4546}.
\newblock


\bibitem[Trewartha et~al\mbox{.}(2022)]%
        {trewartha2022quantifying}
\bibfield{author}{\bibinfo{person}{Amalie Trewartha}, \bibinfo{person}{Nicholas
  Walker}, \bibinfo{person}{Haoyan Huo}, \bibinfo{person}{Sanghoon Lee},
  \bibinfo{person}{Kevin Cruse}, \bibinfo{person}{John Dagdelen},
  \bibinfo{person}{Alexander Dunn}, \bibinfo{person}{Kristin~A Persson},
  \bibinfo{person}{Gerbrand Ceder}, {and} \bibinfo{person}{Anubhav Jain}.}
  \bibinfo{year}{2022}\natexlab{}.
\newblock \showarticletitle{Quantifying the advantage of domain-specific
  pre-training on named entity recognition tasks in materials science}.
\newblock \bibinfo{journal}{\emph{Patterns}} \bibinfo{volume}{3},
  \bibinfo{number}{4} (\bibinfo{year}{2022}), \bibinfo{pages}{100488}.
\newblock
\href{https://doi.org/10.1016/j.patter.2022.100488}{doi:\nolinkurl{10.1016/j.patter.2022.100488}}


\bibitem[Uzzi et~al\mbox{.}(2013)]%
        {uzzi2013atypical}
\bibfield{author}{\bibinfo{person}{Brian Uzzi}, \bibinfo{person}{Satyam
  Mukherjee}, \bibinfo{person}{Michael Stringer}, {and} \bibinfo{person}{Ben
  Jones}.} \bibinfo{year}{2013}\natexlab{}.
\newblock \showarticletitle{Atypical combinations and scientific impact}.
\newblock \bibinfo{journal}{\emph{Science}} \bibinfo{volume}{342},
  \bibinfo{number}{6157} (\bibinfo{year}{2013}), \bibinfo{pages}{468--472}.
\newblock
\href{https://doi.org/10.1126/science.1240474}{doi:\nolinkurl{10.1126/science.1240474}}


\bibitem[Vaswani et~al\mbox{.}(2017)]%
        {vaswani2017attention}
\bibfield{author}{\bibinfo{person}{Ashish Vaswani}, \bibinfo{person}{Noam
  Shazeer}, \bibinfo{person}{Niki Parmar}, \bibinfo{person}{Jakob Uszkoreit},
  \bibinfo{person}{Llion Jones}, \bibinfo{person}{Aidan~N Gomez},
  \bibinfo{person}{\L~ukasz Kaiser}, {and} \bibinfo{person}{Illia Polosukhin}.}
  \bibinfo{year}{2017}\natexlab{}.
\newblock \showarticletitle{Attention is All you Need}. In
  \bibinfo{booktitle}{\emph{Advances in Neural Information Processing
  Systems}}, Vol.~\bibinfo{volume}{30}.
\newblock


\bibitem[Wang et~al\mbox{.}(2023)]%
        {wang2023comprehensive}
\bibfield{author}{\bibinfo{person}{Liyuan Wang}, \bibinfo{person}{Xingxing
  Zhang}, \bibinfo{person}{Hang Su}, {and} \bibinfo{person}{Jun Zhu}.}
  \bibinfo{year}{2023}\natexlab{}.
\newblock \showarticletitle{A Comprehensive Survey of Continual Learning:
  Theory, Method and Application}.
\newblock \bibinfo{journal}{\emph{IEEE Transactions on Pattern Analysis and
  Machine Intelligence}}  \bibinfo{volume}{46} (\bibinfo{year}{2023}),
  \bibinfo{pages}{5362--5383}.
\newblock
\href{https://doi.org/10.1109/TPAMI.2024.3367329}{doi:\nolinkurl{10.1109/TPAMI.2024.3367329}}


\bibitem[Wortsman et~al\mbox{.}(2022)]%
        {wortsman2022model}
\bibfield{author}{\bibinfo{person}{Mitchell Wortsman}, \bibinfo{person}{Gabriel
  Ilharco}, \bibinfo{person}{Samir~Ya Gadre}, \bibinfo{person}{Rebecca
  Roelofs}, \bibinfo{person}{Raphael Gontijo-Lopes}, \bibinfo{person}{Ari~S
  Morcos}, \bibinfo{person}{Hongseok Namkoong}, \bibinfo{person}{Ali Farhadi},
  \bibinfo{person}{Yair Carmon}, \bibinfo{person}{Simon Kornblith},
  {et~al\mbox{.}}} \bibinfo{year}{2022}\natexlab{}.
\newblock \showarticletitle{Model soups: averaging weights of multiple
  fine-tuned models improves accuracy without increasing inference time}. In
  \bibinfo{booktitle}{\emph{International Conference on Machine Learning}}.
  \bibinfo{pages}{23965--23998}.
\newblock


\bibitem[Zhang and Choi(2023)]%
        {zhang2023mitigating}
\bibfield{author}{\bibinfo{person}{Michael Zhang} {and} \bibinfo{person}{Eunsol
  Choi}.} \bibinfo{year}{2023}\natexlab{}.
\newblock \showarticletitle{Mitigating Temporal Misalignment by Discarding
  Outdated Facts}. In \bibinfo{booktitle}{\emph{Proceedings of the 2023
  Conference on Empirical Methods in Natural Language Processing}}.
  \bibinfo{pages}{14213--14226}.
\newblock
\href{https://doi.org/10.18653/v1/2023.emnlp-main.879}{doi:\nolinkurl{10.18653/v1/2023.emnlp-main.879}}


\bibitem[Zhou et~al\mbox{.}(2009)]%
        {zhou2009predicting}
\bibfield{author}{\bibinfo{person}{Tao Zhou}, \bibinfo{person}{Linyuan L{\"u}},
  {and} \bibinfo{person}{Yi-Cheng Zhang}.} \bibinfo{year}{2009}\natexlab{}.
\newblock \showarticletitle{Predicting missing links via local information}.
\newblock \bibinfo{journal}{\emph{The European Physical Journal B}}
  \bibinfo{volume}{71} (\bibinfo{year}{2009}), \bibinfo{pages}{623--630}.
\newblock
\href{https://doi.org/10.1140/epjb/e2009-00335-8}{doi:\nolinkurl{10.1140/epjb/e2009-00335-8}}


\end{thebibliography}
\end{document}